\documentclass[final,5p,times,twocolumn,sort&compress]{elsarticle}
\biboptions{sort&compress}
\usepackage[flushleft]{threeparttable}
\usepackage{graphicx}
\usepackage{makeidx} % allows for indexgeneration
\usepackage{amsmath}
\usepackage{siunitx}
\usepackage{array}
\usepackage{soul}

\usepackage{multirow}% http://ctan.org/pkg/multirow 
\usepackage{longtable}

\usepackage{amsfonts}%
\usepackage{amssymb}%
\usepackage[ruled,vlined]{algorithm2e}
\usepackage{etoolbox}
\makeatletter
\patchcmd{\algocf@Vline}{\vrule}{\vrule\hspace{-0.25em}}{}{}
\makeatother
\usepackage{graphics}
\usepackage{subfig}
\usepackage[switch]{lineno}
\usepackage{float}
\usepackage{setspace}
\makeatletter
\g@addto@macro\endfrontmatter{\enlargethispage{-2\baselineskip}}
\makeatother

\usepackage[labelfont=bf]{caption}
\usepackage[font={small}]{caption}
\usepackage{array}
\usepackage{makecell}

\usepackage{transparent}
\usepackage{color}
\usepackage{xcolor}
\usepackage[utf8]{inputenc}
\usepackage[english]{babel}
\usepackage[hidelinks]{hyperref}

\makeatletter
\newcommand\footnoteref[1]{\protected@xdef\@thefnmark{\ref{#1}}\@footnotemark}
\makeatother
\usepackage{tabularx}
\usepackage{enumitem}
\usepackage{adjustbox}

\usepackage[figuresright]{rotating}

\journal{Imaging Neuroscience}

\begin{document}

\newcommand{\revision}[1]{\textcolor{black}{#1}}
\newcommand{\rev}[1]{\textcolor{black}{#1}}
\begin{frontmatter}

\title{\revision{FastSurfer-HypVINN}: Automated sub-segmentation of the hypothalamus and adjacent structures on high-resolutional brain MRI}
\author[label1,label2]{Santiago Estrada}
\author[label1]{David K\"ugler}
\author[label1,label5]{Emad Bahrami}
\author[label2]{Peng Xu} 
\author[label2]{Dilshad Mousa}  
\author[label2,label4]{Monique M.B.\ Breteler}
\author[label2,label3]{N.\ Ahmad Aziz}
\author[label1,label6,label7]{Martin Reuter\corref{cor1}}

\address[label1]{AI in Medical Imaging, German Center for Neurodegenerative Diseases (DZNE), Bonn, Germany}
\address[label2]{Population Health Sciences, German Center for Neurodegenerative Diseases (DZNE), Bonn, Germany}
\address[label3]{Department of Neurology, Faculty of Medicine, University of Bonn, Bonn, Germany}
\address[label4]{Institute for Medical Biometry, Informatics and Epidemiology (IMBIE), Faculty of Medicine, University of Bonn, Bonn, Germany}
\address[label5]{Computer Science Department, University of Bonn, Bonn, Germany}
\address[label6]{A.A. Martinos Center for Biomedical Imaging, Massachusetts General Hospital, Boston MA, USA }
\address[label7]{Department of Radiology, Harvard Medical School, Boston MA, USA}

\cortext[cor1]{Correspondence to: Martin Reuter (\texttt{martin.reuter [at] dzne.de}).}
\begin{abstract}
The hypothalamus plays a crucial role in the regulation of a broad range of physiological, behavioural, and cognitive functions. However, despite its importance, only a few small-scale neuroimaging studies have investigated its substructures, likely due to the lack of fully automated segmentation tools to address scalability and reproducibility issues of manual segmentation. 
While the only previous attempt to automatically sub-segment the hypothalamus with a neural network showed promise for \SI{1.0}{\milli\meter} isotropic T1-weighted (T1w) MRI, there is a need for an automated tool to sub-segment also high-resolutional (HiRes) MR scans, as they are becoming widely available, and include structural detail also from multi-modal MRI. We, therefore, introduce a novel, fast, and fully automated deep learning method named \revision{\textit{HypVINN}} for sub-segmentation of the hypothalamus and adjacent structures on \SI{0.8}{\milli\meter} isotropic T1w and T2w brain MR images that is robust to missing modalities.
We extensively validate our model with respect to segmentation accuracy, generalizability, in-session test-retest reliability, and sensitivity to replicate hypothalamic volume effects (e.g.\ sex-differences). The proposed method exhibits high segmentation performance both for standalone T1w images as well as for T1w/T2w image pairs.
Even with the additional capability to accept flexible inputs, our model matches or exceeds the performance of state-of-the-art methods with fixed inputs. We, further, demonstrate the generalizability of our method in experiments with \SI{1.0}{\milli\meter} MR scans from both the Rhineland Study and the UK Biobank -- an independent dataset never encountered during training with different acquisition parameters and demographics. \revision{Finally, \textit{HypVINN} can perform the segmentation in less than a minute (GPU) and will be available in the open source \textit{FastSurfer} neuroimaging software suite}, offering a validated, efficient, and scalable solution for evaluating imaging-derived phenotypes of the hypothalamus.

\end{abstract}


\end{frontmatter}

%\linenumbers

\section{Introduction} 

\subsection{Motivation}
The hypothalamus consists of a group of interconnected neuronal nuclei located at the base of the brain~\citep{saper2014hypothalamus}. It is the body’s principal homeostatic center and plays a crucial role in the regulation of a broad range of physiological, behavioural, and cognitive functions, both through direct control of endocrine and autonomic nervous system outflow, as well as through extensive projections to cortical and limbic regions~\citep{saper2014hypothalamus}. Neuropathological studies have demonstrated extensive involvement of the hypothalamus in a range of neurodegenerative diseases, including Alzheimer's disease~\citep{roh2014potential,liguori2014orexinergic}, Parkinson's disease~\citep{fronczek2007hypocretin}, Huntington's disease~\citep{van2021hypothalamic}, frontotemporal dementia, and amyotrophic lateral sclerosis~\citep{bocchetta2015detailed,ahmed2021hypothalamic}. However, the association between hypothalamic integrity and physiological, behavioural, and cognitive outcomes has not been studied in large clinical or population-based studies for lack of a reliable high-throughput automatic imaging procedure.

The majority of studies on hypothalamic imaging-derived phenotypes use manual annotations of magnetic resonance imaging (MRI) scans as the gold standard. Manual segmentation of the hypothalamus and its substructures is commonly done on T1-weighted images~\citep{schindler2013development,makris2013volumetric}. Nonetheless, the use of multi-modal structural information during the manual annotation process has also been proposed to increase especially the visibility of the lateral hypothalamus boundaries~\citep{bocchetta2015detailed,baroncini2012mri}. These multi-modal protocols recommend segmenting the hypothalamus using simultaneous visualization of registered T1-weighted (T1w) and T2-weighted (T2w) MR images. Manual delineation of the hypothalamus, however, is a very time-consuming process that relies highly on the user's expertise due to the small size and low boundary MR contrast in the hypothalamus region, regardless of the available MRI modalities.

Automated methods have been proposed to segment the whole \revision{hypothalamus~\citep{orbes2015magnetic,thomas2019higher,rodrigues2020hypothalamus,rodrigues2022benchmark,greve2021deep}} and its sub-regions~\citep{billot2020automated} quickly and reliably. However, even though automated tools are available, they only focus on segmenting \SI{1.0}{\milli\meter} isotropic T1w scans, ignoring the detailed structural information available in sub-millimeter resolution datasets. High-resolutional (HiRes) MR scans are becoming more common across studies (even in clinical settings) due to rapid advancements in MR technology (e.g.\ accelerated acquisition schemes) and are increasingly employed as the new standard for large studies (e.g.\ the Rhineland Study~\citep{rs1-stocker2016big,rs2-breteler2014mri}, Human Connectome Project (HCP) datasets~\citep{hcp-van2012human,bookheimer2019lifespan,harms2018extending}, Autism Brain Imaging Data Exchange II (ABIDE-II)~\citep{abide-di2017enhancing}, TRACK-PD~\citep{wolters2020track}). Thus, the need for neuroimaging tools that can handle sub-millimeter resolutions (e.g.\ \SI{0.8}{\milli\meter} isotropic) has increased. 

\revision{Moreover, current automated hypothalamic segmentation methods have neglected the inclusion of multi-modal structural information}. One reason for this is that simultaneous access to T1w and T2w images is not always possible due to constraints in scanning time or poor image quality in one of the modalities due to reduced image resolution or acquisition artefacts. Therefore, the introduction of an accurate automated method for segmenting hypothalamic structures on high-resolutional T1w and T2w MRI scans, which is also robust to handle missing modalities, is of significant interest to clinicians and researchers.

\subsection{Related work}
Automated hypothalamic segmentation methods utilizing multi-atlas-based techniques~\citep{orbes2015magnetic,thomas2019higher} were initially proposed. However, these methods are slow and demand considerable computational resources. Newer techniques such as fully convolutional neural networks (F-CNNs) can tremendously speed-up computation time by utilizing graphical processing units (GPUs) and have become the preferred method for solving supervised semantic segmentation problems in the medical computer vision community~\citep{henschel2020fastsurfer,kamnitsas2017efficient,estrada2020fatsegnet,quicknat,ronneberger2015u,milletari2016fully,estrada2021automated,faber2022cerebnet}.  

\revision{Hypothalamus segmentation using F-CNNs has mainly focused on identifying the hypothalamus as one whole structure in the  brain~\citep{rodrigues2020hypothalamus,rodrigues2022benchmark,greve2021deep}}. Recently, \textit{Billot et al.}~\citep{billot2020automated} proposed a method to segment five sub-regions of the hypothalamus using an encoder-decoder 3D F-CNN with extensive data augmentation. They followed the hypothalamic parcellation protocol introduced by \textit{Makris et al.}~\citep{makris2013volumetric} on standard \SI{1.0}{\milli\meter} isotropic resolution T1w images. Their proposed method illustrates the capabilities of F-CNNs to segment hypothalamic compartments with promising results on datasets acquired at \SI{1.0}{\milli\meter} isotropic resolution~\citep{billot2020automated,shapiro2022vivo}. However, F-CNNs are known to have issues generalizing to resolutions that differ from the training one~\citep{estrada2021automated,iglesias2021joint,henschel2022fastsurfervinn} rendering HiRes images out-of-distribution and unsuitable for methods designed for lower resolutions. A common approach for this problem is to down-sample the input image to the desired lower resolution in a \revision{pre-processing step~\citep{henschel2020fastsurfer,billot2020automated,greve2021deep}}. This process, however, reduces image details and information, forfeiting the investment already made when acquiring the higher resolution in the first place. Furthermore, HiRes information could help address inter-class inconsistencies between voxels at a local and global level and alleviate the partial volume effect problem~\citep{glasser2013minimal}. 

HiRes segmentation of brain structures has mostly been tackled by training with manual annotations created at the desired resolution~\citep{estrada2021automated,rushmore2022anatomically,kamnitsas2017efficient,beliveau2021automated} or training models using \SI{1.0}{\milli\meter} data with scale-augmentations -- an established deep-learning technique to improve the generalizability of a model. \revision{Recently, models capable of segmenting scans at different resolutions have been introduced. \textit{Billot et al.}~\citep{billot_synthseg_2023,billot_robust_2023} proposed \textit{SynthSeg}, a technique for generating segmentations at a fixed resolution (\SI{1.0}{\milli\meter}), regardless of the resolution of the input scan, which are interpolated to the fixed resolution as a pre-processing step. During training, \textit{SynthSeg} relies on a generative model that produces ``unrealistic synthetic images''~\citep{billot_synthseg_2023}. These synthetic images are created from ground truth label maps at the pre-defined fixed resolution. This approach simulates domain variability by incorporating multiple random parameters for the generator, such as spatial, intensity, contrast, and resolution variability. While providing input flexibility, the model's output resolution, however, remains confined to the fixed resolution}. 

\revision{Before \textit{SynthSeg}, we introduced the Voxel-Size Independent Neural Network (\textit{VINN}) for resolution-independent segmentation tasks~\citep{henschel2022fastsurfervinn}. The VINN approach enables training and inference using images at multiple resolutions within a single network. In brief, instead of interpolating input images, \textit{VINN} integrates the resolution change into the network, replacing a regular scale transition with an interpolation layer, that maps the latent space at native input resolution to a pre-defined internal resolution at lower layers of the network and vice-versa. 
As a result, rich HiRes information is retained without image or label interpolation, and segmentations are provided at the desired native input resolution}.   

\revision{Finally, as has already been shown in manual segmentation of hypothalamic structures, exclusively utilizing T1w images as input forfeits the significant potential presented by the inclusion of multi-modal information (T1w and T2w)~\citep{bocchetta2015detailed,baroncini2012mri}}. Common multi-modal F-CNN architectures, however, require all input modalities to always be present. The absence of any modality introduces a computational bias that the network is not trained to handle. To overcome missing modalities, \revision{proposed solutions include} training a specific network for each of the input combinations or providing the segmentation model with a synthesized version of unavailable modalities~\citep{van2015does,hofmann2008mri}. \revision{Alternatively, training networks with synthetic image contrast has also been suggested~\citep{billot_synthseg_2023, billot2020learning}}. Even though these techniques have shown promising results, a more suitable model should be capable of extracting the most salient information for solving the given task from the available modalities without the need for artificial images or multiple modality-specific networks. With this in mind, shared latent space models were introduced on the challenging task of multi-modal brain tumor segmentation~\citep{hemis_havaei2016,pimms_varsavsky2018,hetero_variational_dorent2019}. This approach first translates modalities into independent latent spaces; afterwards, the modalities' embedded information is merged inside the network into a shared latent representation. The shared latent space is then forwarded to the remaining network to solve the desired task. At inference time, the shared representation is computed from the available modalities, thus being robust to all input-modality combinations (i.e.\ hetero-modal) included in training.
 
To address the missing modalities challenge \revision{in a HiRes scenario}, we suitably include the shared latent space concept into our voxel-size independent network \revision{(\textit{VINN}). \textit{Hetero-modal VINN} (\textit{HM-VINN})} introduces a fusion module that linearly combines the modalities inside the network. After passing the available scans through a separate modality-specific convolutional block, the network weighs and merges the feature maps based on the best available information using a learnable weighted sum. As the output of the fusion module is normalized, missing one modality can be tackled by assigning zero to its respective weight.

\begin{figure*}[!hbt]
    \centering
    \includegraphics[width=0.95\textwidth]{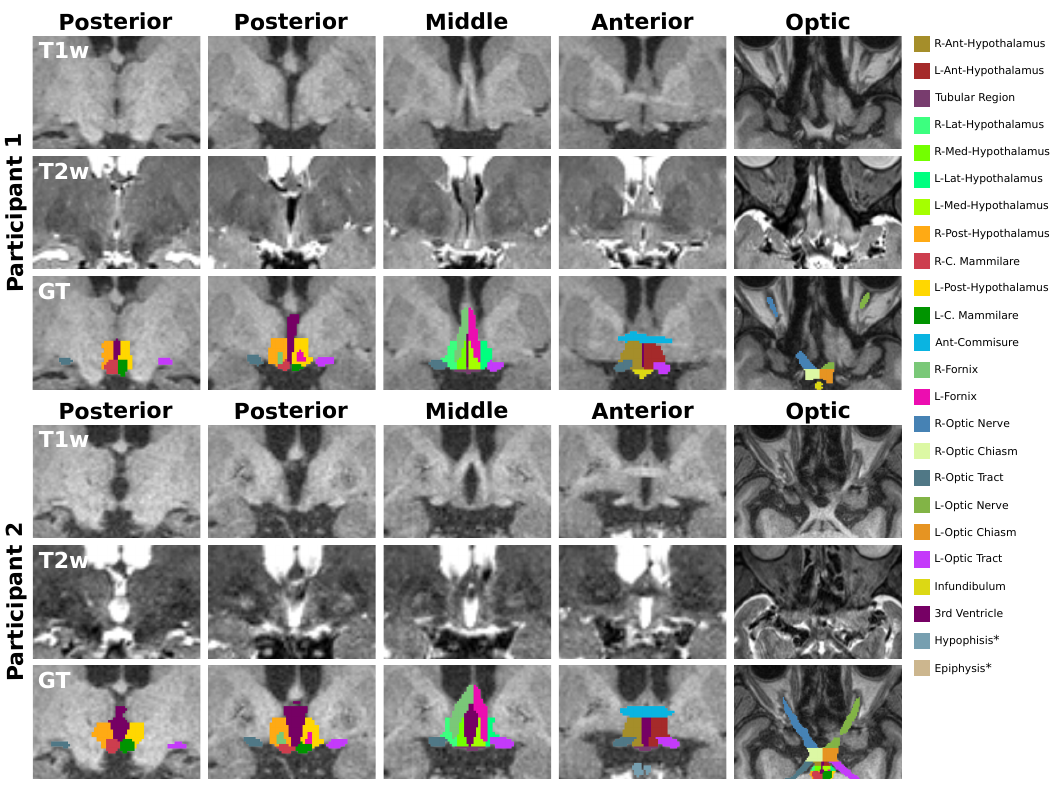}
        \caption{T1-weighted (T1w) and T2-weighted (T2w) images and ground truth (GT) from two participants. The proposed manual segmentation scheme is composed of twenty-four structures divided into three major regions: 1) hypothalamic (anterior, middle, and posterior), 2) optic, and 3) others. The color lookup table\textsuperscript{*} for all structures is presented on the left, and a detailed overview of the three regions is presented in Table~\ref{tab:gt_labels}. \textsuperscript{*}~Structures are not visible in the presented snapshots.}
    \label{fig:gt_example}
\end{figure*}

\subsection{Contribution}

To our knowledge, we are the first to tackle automated hetero-modal sub-segmentation of the hypothalamus and adjacent structures on high-resolutional brain MRI. The contributions of this work are the following: Firstly, we introduce a new hypothalamic labeling protocol adapted to the higher spatial resolution offered by 3T \SI{0.8}{\milli\meter} isotropic MR images. The proposed protocol presents a more fine-grained parcellation of the hypothalamus and includes usually ignored brain structures, such as hypophysis, epiphysis, the optic nerve, optic chiasm, and optic tract, as illustrated in Figure~\ref{fig:gt_example}. \revision{Secondly, we present \textit{HypVINN}, a novel automated hypothalamic parcellation tool with a novel \textit{hetero-modal VINN (HM-VINN)} architecture at its core, providing a solution to the multi-resolution and the missing modality challenge in a single model}. We extensively show that the model's input flexibility does not compromise performance compared to state-of-the-art methods with fixed inputs in terms of segmentation accuracy, test-retest reliability, and generalizability. \revision{Moreover, our method replicates hypothalamic volume effects (e.g.\ age and sex) on subsets of the \SI{0.8}{\milli\meter} (HiRes) Rhineland Study (n=463) and the \SI{1.0}{\milli\meter} UK Biobank (n=535)~\citep{ukb1-miller2016multimodal,ukb2-alfaro2018image}.} Last but not least, and to the benefit of the research community, we will integrate the \revision{\textit{HypVINN}} tool into the user-friendly, open source \textit{FastSurfer} framework~\citep{henschel2020fastsurfer} available at: https://github.com/Deep-MI/FastSurfer (code will be released upon acceptance).

\section{Methods}

\subsection{Datasets} \label{sec:mri_data}

We used MR images from two population studies, namely the Rhineland Study (RS) ~\citep{rs1-stocker2016big,rs2-breteler2014mri} and the UK Biobank (UKB)~\citep{ukb1-miller2016multimodal,ukb2-alfaro2018image}, with resolutions of \SI{0.8}{\milli\meter} (HiRes) and \SI{1.0}{\milli\meter}, respectively. Participants from both studies gave written informed consent in accordance with the ethical guidelines of the individual studies. Furthermore, ethics approval and regulations can be accessed on their respective webpages. For this work, we compiled four distinct datasets from the population studies: a manually annotated dataset (from RS), a generalizability dataset (from RS and UKB), a test-retest dataset (from RS), and a \revision{case-study dataset (from RS and UKB)}. The manually annotated dataset (referred to as "in-house dataset") was initially split into two non-overlapping sets, one for training and validation, and the other for testing. The remaining datasets were exclusively used for evaluations to assess different aspects of our hetero-modal method.

The Rhineland Study is an ongoing population-based cohort study located in Bonn, Germany, which enrolls participants aged 30 years and above (www.rheinland-studie.de). MR scans were collected at two different sites using identical 3T Siemens MAGNETOM Prisma MRI scanners equipped with 64-channel head-neck coils. The core MRI acquisition protocol for every participant in the Rhineland Study includes the following MR contrast: T1w, T2w, FLAIR, diffusion-weighted, susceptibility-weighted, resting-state functional, and abdominal Dixon MRI with a total net scan time of around 45 minutes. Furthermore, an optional extra acquisition time (maximum 10 minutes) is available for a free protocol. 

This paper utilized the \SI{0.8}{\milli\meter} isotropic T1w and T2w MR scans. The T1 protocol consists of a multi-echo magnetization prepared rapid gradient echo (MPRAGE) sequence~\citep{van2008brain} with 2D acceleration~\citep{brenner2014two}, while the T2 protocol uses a 3D Turbo-Spin-Echo (TSE) sequence with variable flip angles~\citep{busse2008effects}. Both sequences also utilize elliptical sampling~\citep{mugler2014optimized} and parallel imaging (PI)~\citep{griswold2002generalized} to expedite the imaging process. For this work, all protocol versions from the Rhineland Study were considered, and sequence parameters are presented in Appendix Table~\ref{tab:t1_t2_parameters}. 

We compiled the Rhineland Study datasets by first randomly selecting a subset (n=534) of participants with available T1w and T2w scans from sex and age strata to ensure a balanced population distribution. The sample presents a mean age of 54.9 years (range 30 to 95), and 59.4\% were women. We then further assigned participants to the in-house dataset and all its subsequent splits adhering to the age and sex-stratification scheme. All T2w scans were registered to their corresponding T1w scan using \textit{FreeSurfer}'s mri\_robust\_register tool~\citep{reuter2010highly}.

MRI scans of the \textit{in-house training and testing dataset} (n=50) were manually annotated by an experienced rater and split into training/validation (n=44) and testing (n=6) sets. Training data was further split into four groups for cross-validation. Finally, the testing data was manually annotated for a second time by our main rater to evaluate intra-rater variability. The rater was blind to the scans' identification to avoid bias and overestimating performance. 

For evaluating within-session \textit{test-retest} reliability, we utilized the RS subset (n=21) with two in-session T1w scans. The additional scan for this participant was acquired during the time slot allocated for a free protocol inside the Rhineland study's MRI acquisition protocol. Due to the time constraint of the free protocol, a second T2w scan could not be acquired. Before starting the free protocol, participants were asked to move their head inside the head-neck coil. It is important to note, that T1w scans were not acquired back-to-back, but with a time gap of almost 30 minutes. 

The MRI scans of the remaining participants (n=463) were compiled into the \revision{\textit{RS case-study dataset}} to evaluate the sensitivity to known hypothalamic volume effects (e.g.\ age and sex). For a detailed description of the population characteristics of all the aforementioned \revision{RS} subsets see Appendix Table~\ref{tab_appendix:rs_demo_table} and~\ref{tab_appendix:rs_train_demo_table}.

We used data from the UK Biobank study to test the \textit{generalizability} of our method to isotropic \SI{1.0}{\milli\meter} scans from an unseen cohort with different acquisition parameters. \revision{An initial subset (n=544) of random participants was selected from sex and age strata to ensure a balanced population distribution. The chosen sample presents a mean age of 58.7 years (range 45 to 82), consisting of 52.6\% women. Subsequently, the scans of nine random participants were manually labeled by our expert rater to evaluate segmentation accuracy at \SI{1.0}{\milli\meter} (\textit{generalizability dataset}). The remaining UKB participants (n=535, \textit{UKB case-study dataset}) were also used in the hypothalamic volumes effects sensitivity analysis. A summary of the population characteristic of the UKB subsets is presented in Appendix Table~\ref{tab_appendix:ukb_demo_table}}.

\subsection{Manual reference standard}\label{sec:labels}
An experienced rater manually annotated the sub-regions of the hypothalamus and adjacent structures on registered T1w and T2w images, except for the UK Biobank cases where only T1w scans were available. The annotation was performed using \textit{Freeview}, a visualization tool of \textit{FreeSurfer}~\citep{freesurfer1,freesurfer2}, which allowed simultaneous viewing of the available modalities. Summarizing the labeling process, the borders of the unilateral hypothalamus were defined as follows~\citep{makris2013volumetric}: a) anteriorly: coronal plane passing through the most rostral tip of the anterior commissure and containing the optic chiasm, b) posteriorly: coronal plane through the most caudal tip of the mammillary bodies, c) superiorly: third ventricle with the diencephalic fissure, d) inferiorly: junction to the optic chiasm rostrally and the hemispheric margin more caudally, e) medially: wall of the third ventricle and the interhemispheric fissure, and f) laterally: rostrally at the medial border of the optic tract and more caudally at the internal capsule, globus pallidus and cereberal penduncle. A detailed definition of the segmentation procedure for all different substructures is provided in~\ref{sec_appendix:manual}. Adjacent small hypothalamic nuclei were grouped into subunits according to Table~\ref{tab:gt_labels}. An example of the manual segmentation scheme is illustrated in Figure~\ref{fig:gt_example}, and an overview of all twenty-four segmented structures is presented in Table~\ref{tab:gt_labels}.

\begin{table*}[!hbt]
\centering
\caption{Summary of the hypothalamic sub-regions and adjacent structures included in the proposed labeling scheme with its corresponding name, anatomical designation and region.}
\label{tab:gt_labels}
\resizebox{0.95\textwidth}{!}{%
\renewcommand{\arraystretch}{1.2}% Default value: 1
\begin{threeparttable}
\begin{tabular}{lp{0.3\textwidth}lllll}
\hline
\multicolumn{3}{l}{Hypothalamic Sub-regions}                                                            & & \multicolumn{3}{l}{Adjacent Structures}                                             \\ \cline{1-3} \cline{5-7} 
Label Name                          & Anatomical designation                      & Region Group               & & Label Name        & Anatomical designation         & Region Group           \\ \hline
\multirow{2}{*}{L-Ant-Hypothalamus} & \multirow{2}{0.3\textwidth}{Anterior Hypothalamus (lh), Supraoptic Nucleus (lh)}   
                                                                                   & \multirow{4}{*}{Anterior}  & & 3rd-Ventricle     & 3rd-Ventricle, Superior-Border & \multirow{7}{*}{Others} \\
                                    &                                              &                            & & L-Fornix          & Fornix (lh)                    &                          \\  
\multirow{2}{*}{R-Ant-Hypothalamus} & \multirow{2}{0.3\textwidth}{Anterior Hypothalamus (rh), Supraoptic Nucleus (rh)}                             
                                                                                   &                            & & R-Fornix          & Fornix (rh)                    &                          \\ 
                                    &                                              &                            & & Epiphysis         & Epiphysis                      &                          \\ \cline{1-3}
L-Med-Hypothalamus                  & Medial Hypothalamus\textsuperscript{*} (lh)  & \multirow{7}{*}{Middle}    & & Hypophysis        & Hypophysis, Neurohypophysis    &                          \\
R-Med-Hypothalamus                  & Medial Hypothalamus\textsuperscript{*} (rh)  &                            & & Infundibulum      & Infundibulum                   &                        \\
L-Lat-Hypothalamus                  & Lateral-Hypothalamus (lh)                    &                            & & Ant-Commisure     & Anterior Commisure            &                          \\\cline{5-7} 
R-Lat-Hypothalamus                  & Lateral-Hypothalamus (rh)                    &                            & & L-N-Opticus       & Optic Nerve (lh)               & \multirow{6}{*}{Optic}   \\
\multirow{3}{*}{Tuberal-region}     & \multirow{3}{0.3\textwidth}{Median-eminence, Tuberomammillary Region, and Arcuate-nucleus} 
                                                                                   &                            & &  R-N-Opticus      & Optic Nerve (rh)               &                          \\
                                    &                                              &                            & & L-Chiasma-Opticus & Optic Chiasm (lh)              &                             \\ 
                                    &                                              &                            & & R-Chiasma-Opticus & Optic Chiasm (lh)              &                             \\ \cline{1-3}
L-Post-Hypothalamus                 & Posterior Hypothalamus (lh)                  & \multirow{4}{*}{Posterior} & & L-Optic-tract     & Optic Tract (lh)               &                              \\ 
R-Post-Hypothalamus                 & Posterior Hypothalamus (rh)                  &                            & & R-Optic-tract     & Optic Tract (rh)               &                             \\ 
L-C-Mammilare                       & Corpus Mammillare (lh)                       &                            & &                   &                                &                             \\
R-C-Mammilare                       & Corpus Mammillare (rh)                       &                            & &                   &                                &                          \\ \hline
\end{tabular}
\begin{tablenotes}  
\item \textsuperscript{*} including the Paraventricular Nucleus (PVN), the Ventromedial Nucleus (VMN), and the Dorsomedial Nucleus (DMN).
\end{tablenotes}
\end{threeparttable}}
\end{table*}

\subsection{\revision{Hypothalamic hetero-modal segmentation tool - \textit{HypVINN}}}

\subsubsection{Hetero-modal segmentation network - \revision{\textit{HM-VINN}}} \label{sec:network}

\begin{figure*}[!hbt]
    \centering
    \includegraphics[width=0.95\textwidth]{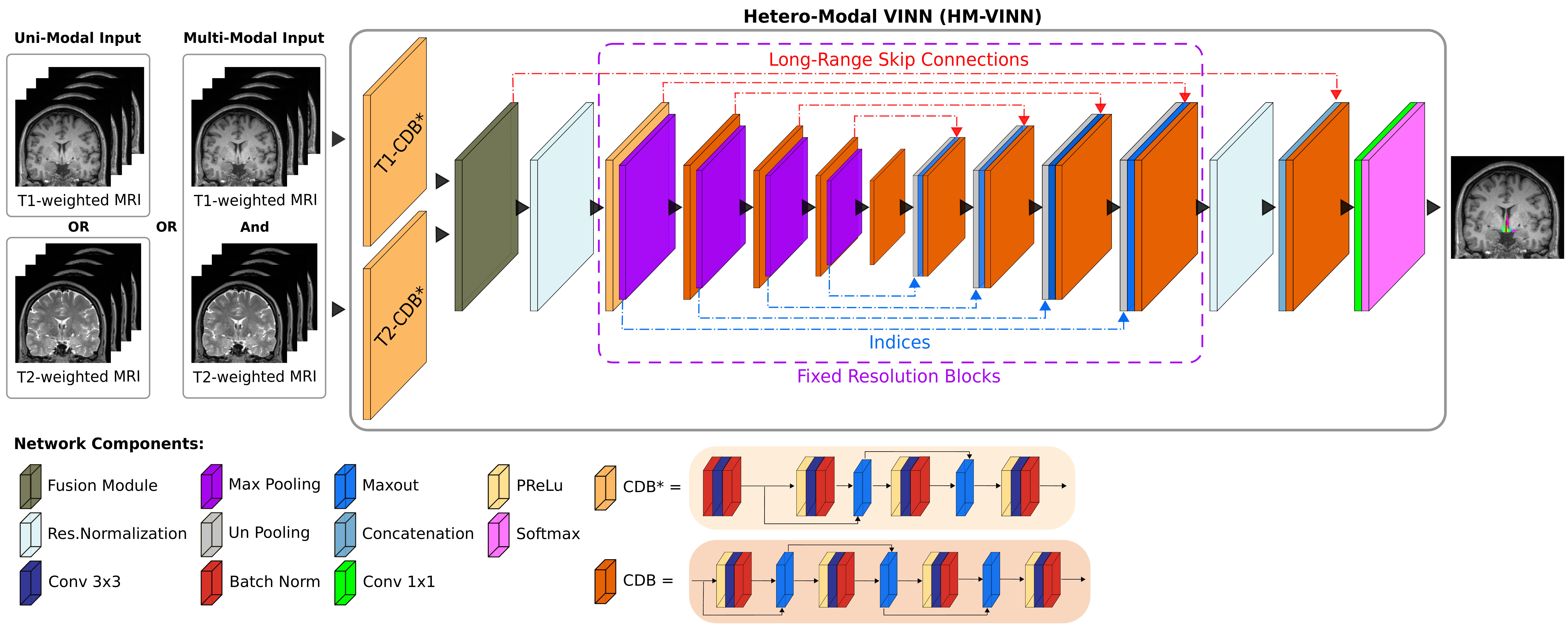}
     \caption{\revision{\textit{Hetero-Modal VINN (HM-VINN)} architecture in \textit{HypVINN}}. Input modalities are first independently processed by modality-specific competitive dense blocks (T1-CDB\textsuperscript{*} and T2-CDB\textsuperscript{*}). Afterward, modality-specific feature maps are merged inside the network by our proposed fusion module \rev{(dark green)} to create a shared latent space. During inference time, the shared latent space can be computed over the available modalities and fed into the remaining network. Furthermore, \revision{\textit{HM-VINN}} incorporates flexible transitions in the first and last scale transition by utilizing the network-integrated resolution-normalization \rev{(light blue)}. Each CDB is composed of four sequences of parametric rectified linear unit (PReLU), convolution (Conv), and batch normalization (BN). In the modality-specific CDBs and second encoder block (CBD\textsuperscript{*}), the first PReLU is replaced with a BN to normalize the inputs.}
    \label{fig:network_schematic}
\end{figure*}

To accurately segment the hypothalamic sub-regions and adjacent structures, we employ \revision{\textit{VINN}}~\citep{henschel2022fastsurfervinn} as the foundation for our network design. \revision{\textit{VINN}} is a resolution-independent extension of the successful multi-network approach \textit{FastSurferCNN}~\citep{henschel2020fastsurfer,estrada2021automated,faber2022cerebnet}. Both methods are 2.5D approaches, i.e.\ they aggregate predictions of three 2D F-CNNs (one per anatomical view) with multi-slice input~\citep{henschel2020fastsurfer}. The F-CNNs follow a UNet-type layout with an encoder and decoder arm of five competitive-dense blocks (CDB) separated by an additional bottleneck CDB (see Figure~\ref{fig:network_schematic}). In \textit{FastSurferCNN}, all scale transitions between the CDBs are implemented via fixed-scale down- or up-sampling operations (i.e.\ (un)pooling). \revision{\textit{VINN}}, on the other hand, replaces the first and last scale transition with a flexible network-integrated resolution-normalization. Here, the native image resolution is explicitly integrated into the network and utilized to interpolate the feature maps to a common pre-defined network base resolution (\SI{1.0}{\milli\meter}). In turn, network capacity in the inner layers is available for the segmentation task while retaining voxel size-dependent information outside of it. Lastly, the view-aggregation step ensembles the resulting probabilities maps through a weighted average (axial = 0.4, coronal = 0.4, and sagittal = 0.2). The weights of the sagittal predictions are reduced compared to the other predictions, as structures with left and right hemispheres labels are unified into one due to missing lateralization information in the sagittal view~\citep{henschel2020fastsurfer}. For the current segmentation task, we also unify lateralized structure labels into one for the sagittal view, consequently reducing the number of classes in the sagittal F-CNN from 24 to 15. Therefore, the \revision{\textit{VINN}} view-aggregation weighting scheme is also suitable for our application.  

\revision{In this work, we extend \revision{\textit{VINN}} into a hetero-modal segmentation scenario (referred to as \revision{\textit{HM-VINN}}) by embedding the input modalities into a shared latent space~\citep{hemis_havaei2016,pimms_varsavsky2018,hetero_variational_dorent2019}}. Following this direction, we modify the standard F-CNNs from \textit{VINN} to initially process T1w and T2w images independently of each other by replacing the first encoder CDB with modality-specific CDBs (Figure~\ref{fig:network_schematic}, e.g.\ T1-CDB\textsuperscript{*} and T2-CDB\textsuperscript{*}). After the independent stage, feature maps are merged inside the network by a fusion module and fed into the following convolutional pipeline.

The implemented fusion module weights and merges the feature maps from the T1 and T2 branches based on the best available information using a learnable weighted sum. Let us denote the output feature map from the T1-CDB\textsuperscript{*}  as $F_{T1}\:\epsilon\:\mathbb{R}^{C \times H \times W}$ and the T2-CDB\textsuperscript{*}  output as $F_{T2}\:\epsilon\:\mathbb{R}^{C \times H \times W}$, where $C,H,W$ represent the channel, height, and width dimensions, respectively. Then, the output of fusion module ($F_\text{fused}$) is 
\begin{linenomath*}
 \begin{equation}
\begin{aligned}
F_\text{fused} = \frac{\left | W_{T1} \right |}{\left | W_{T1} \right | + \left | W_{T2} \right |}\times F_{T1} + \frac{\left | W_{T2} \right |}{\left | W_{T1} \right | + \left | W_{T2} \right |}\times F_{T2} \text{ ,}
\label{eq:1}
\end{aligned}
\end{equation}
\end{linenomath*}
where $W_{T1}$ and $W_{T2}$ are global learnable scalar parameters initialized both at 0.5. The introduction of $W_{T1}$ and $W_{T2}$ allows the network to gradually learn the importance of each modality. If a modality is more informative, its feature maps will have a higher weight. Additionally, as the output of the fusion module is normalized, missing one modality can be tackled by assigning zero to its respective weight. Thus, the fusion features are identical to the encoder block output of the existing modality. 
 
In detail, all three F-CNNs followed the abovementioned layout (see Figure~\ref{fig:network_schematic}). Within F-CNNs, the CDB layout is kept mostly the same as the one from \textit{VINN}, where the CDB consists of four layers of parametric rectified linear unit (PReLU), convolution (Conv - kernel size of $3\times3$), and batch normalization (BN) except for the first two encoders blocks. In the first two encoder blocks from \revision{\textit{VINN}}, the first PReLU is replaced with a BN to normalize the inputs (see Figure~\ref{fig:network_schematic}, CBD\textsuperscript{*}). The modified CBD construction is also utilized for modality-specific CDBs as they are our initial first encoder CDB. To keep the comparison fair in light of an effective parameter count of approx.\ 4.5M parameters (three dedicated, modality-specific models with approx.\ 1.5M parameters each), we increase the number of channels (features) of all layers from 64 to 80 inside CDBs,  and from 32 to 64 in the first and last CDB blocks (i.e.\ the first scale level). This change raises the parameter count to approx.\ 2.6M, which is still significantly less than three ($\approx$~ 4.5M parameters) or even two ($\approx$~3.0M) dedicated, modality-specific networks.

\subsubsection{Hetero-modal training procedure} \label{sec:hetero_learning}
Introducing additional variations by data augmentation during training helps neural networks to be more robust. Here, we make \revision{\textit{HM-VINN}} robust to missing modalities by sometimes randomly dropping either the T1w or T2w image for a given training example with a uniform distribution between all input combinations (modality dropout). The modality weights in the fusion module are adjusted as follows: i) When the two modalities are available, the network automatically assigns the weights (see Eq.\ \ref{eq:1}). ii) If a modality is dropped, its corresponding fusion weight is set to zero as described in the previous section. By starting this modality dropout procedure only after ten epochs, the proposed training procedure first establishes general segmentation capabilities (with all modalities available) before pivoting to more difficult scenarios with different combinations and missing modalities.

\subsubsection{Model learning} \label{sec:model_learning} 
All F-CNN are implemented in PyTorch~\citep{paszke2017automatic} using a docker container~\citep{merkel2014docker}. Independent models for axial, coronal, and sagittal views are trained for 100 epochs with batch size of 16 using two NVIDIA Tesla V100 GPU with 32 GB RAM. We use the AdamW~\citep{kingma2015adam, loshchilov2018decoupled} optimizer with a weight decay of $10^{-4}$ and an initial learning rate of $0.05$, which is decreased to $0.005$ after 70 epochs.  The networks are trained by optimizing a combined loss function of a median frequency weighted cross-entropy loss and Dice loss~\citep{quicknat}. This loss function encourages correct segmentation along anatomical boundaries and counters class imbalances by increasing the  weights of less frequent classes.

To increase the generalizability of our model, we apply several spatial and intensity data augmentations during training. Spatial augmentations on the inputs images are limited to random affine transformations such as translation (range: from \SIrange{-15}{15}{\milli\meter}), rotation (range:  from \ang{-10} to \ang{10}), and uniform scaling (factor: from 0.85 to 1.15) \citep{perez-garcia_torchio_2020}. 
Furthermore, we include internal scale augmentations of the feature maps as introduced by \textit{FastSurferVINN} to improve the segmentation performance~\citep{henschel2022fastsurfervinn}.

Intensity augmentations are carried out to address two challenges: 1) intensity inhomogeneities due to scan parameters~\citep{perez-garcia_torchio_2020} and 2) artefacts introduced by defacing algorithms in regions of interest (e.g.\ optic region). The first problem is tackled by applying a random bias field~\citep{VanLeemput1999,sudre2017longitudinal} transformation on the input images (coefficients range: from -0.5 to 0.5). For the second issue,  we improve the network's robustness to handle defaced scans by including scans with or without face features as part of the training set. For creating the modified scans, three common open-source algorithms are used (PyDeface~\citep{pydeface}, MiDeFace from \textit{FreeSurfer}~\citep{freesurfer2}, and HCP face masking~\citep{facemasking_milchenko2013obscuring}). In contrast to all above-mentioned transformations, defacing is performed statically before training ("offline") due to the high computation time to deface a scan (more than 1 minute per method).

\subsection{Evaluation metrics}
We compute three standard segmentation metrics (dice similarity coefficient, volume similarity, and Hausdorff distance) to assess the similarity between the predicted label maps and manual annotations~\citep{taha2015metrics}. We first evaluate the dice similarity coefficient~(Dice)~\citep{sorensen1948method,dice1945measures} as it provides spatial overlap consensus. Let M (manual annotations) and P (prediction) denote binary label maps, then Dice is defined as:
\begin{linenomath*}
\begin{equation}
\begin{aligned}
         Dice=  \frac{2\cdot \left | M \cap P\right | } { \left | M\right | + \left |  P\right|}
\end{aligned}
\end{equation}
\end{linenomath*}
where  $| M \cap P |$ represent the number of common elements (intersection) and $|M|$ and $|P|$ the number of elements in each label map, therefore, Dice values range from 0 to 1, and a higher Dice represents a better segmentation agreement. Afterwards, we compute volume similarity~(VS) as volume measurements are usually the desired image-derived phenotype for downstream statistical analysis. VS is defined as:
\begin{linenomath*}
\begin{equation}
\begin{aligned}
        VS = 1 - \frac{\left | |M| - |P| \right |}{|M| + |P| }  .
\end{aligned}
\end{equation}
\end{linenomath*}
VS has the same range as Dice; however, it can have its maximum value even when the spatial overlap is zero, as this metric does not consider spatial localization information. Additionally, a spatial distance-based metric is used to evaluate the quality of segmentation boundary delineation (contour). Here, we use the 95\% Hausdorff distance~(HD95), a Hausdorff distance~(HD) as it is less sensitive to outliers~\citep{huttenlocher1993comparing}. HD95 is considered as the 95th percentile of the ordered distance measures, and it is defined as:
\begin{linenomath*}
\begin{equation}
\begin{aligned}
d_{95}(M,P) =  95^{th}_{m \in M}\left (  \min_{p \in P} d(m,p) \right ) \\
d_{HD95}(M,P) = \max(d_{95}(M,P),d_{95}(P,M))
\end{aligned}
\end{equation}
\end{linenomath*}
where $d$ is the Euclidean distance. In contrast to the Dice and VS,  HD95 is a dissimilarity metric so a smaller value indicates a better boundary delineation with a value of zero being the minimum (perfect match). 

Finally, statistical significant differences in segmentation performance are confirmed throughout this work by a non-parametric paired two-sided Wilcoxon signed-rank test~\citep{wilcoxon1992individual} after correcting for multiple testing using Bonferroni correction (referred to as corrected $p$).

For accessing the test-retest reliability of predicted volume measurements between two repeated scans of the same participant, we use the intra-class correlation (ICC). The ICC is a commonly used metric to assess the degree of agreement and correlation between measurements. The ICC values range from 0 to 1, with higher values representing better reliability. Here, we compute a two-way fixed, absolute agreement and single measures with a 95\% confidence interval (ICC(A,1))~\citep{mcgraw1996forming}.

\section{Experiments and results}

\begin{table*}[!hbt]
\centering
\caption{Mean (and standard deviation) segmentation performance of the cross-validated  F-CNN models on the unseen test-set. The proposed hetero-modal \revision{\textit{HypVINN}} performs as well as the modality-specific models. Furthermore, \revision{\textit{HypVINN}} with multi-modal and standalone T1w input outperforms the 3D-UNet proposed by \textit{Billot et al.}~\citep{billot2020automated} -- the only other contemporary method for hypothalamus parcellation. Note: the statistical significance column (Signif.) indicates, which other models the model outperforms (paired Wilcoxon signed-rank test, corrected $p < 0.05$).}
\label{tab:accuracy_cv_test_set}
\resizebox{0.95\textwidth}{!}{%
\renewcommand{\arraystretch}{1.1} % Default value: 1
\begin{tabular}{llllllllll}
\hline
\multicolumn{1}{l}{\textbf{Experimental Setup}}          & & \multicolumn{2}{l}{\textbf{Dice} \(\uparrow\)}        & & \multicolumn{2}{l}{\textbf{VS} \(\uparrow\)}      & & \multicolumn{2}{l}{\textbf{HD95 (mm)} \(\downarrow\)}             \\ 
\cline{1-1}\cline{3-4} \cline{6-7} \cline{9-10}
Model                                  & & Mean (SD)                & Signif.               & & Mean (SD)                & Signif.      & & Mean (SD)                & Signif.                 \\ \hline
\multicolumn{10}{l}{\textbf{Only T1w input }} \\
a: \revision{T1-VINN} \citep{henschel2022fastsurfervinn}                         & & 0.7937 (0.0926)          & $^{c,d,e}$            & & 0.9030 (0.0785)          & $^{c,e}$     & & 1.1262 (0.5443)          & $^{c,d,e}$            \\
b: \revision{HypVINN}  (Ours)                         & & 0.7905 (0.0968)          & $^{c,d,e}$            & & 0.9053 (0.0757)          & $^{c,d,e}$   & & 1.1312 (0.5683)          & $^{c,d,e}$                \\ 
c: 3D-UNET~\citep{billot2020automated} & & 0.7481 (0.1516)          & $^{d,e}$              & & 0.8753 (0.1325)         & $^{e}$       & & 1.4088 (2.235)           & $^{e}$                           \\ \hline
\multicolumn{10}{l}{\textbf{Only T2w input }} \\
d: \revision{T2-VINN} \citep{henschel2022fastsurfervinn}                         & & 0.7457 (0.1059)          & $^{e}$                & & 0.8967 (0.0877)          & $^{c,e}$     & & 1.2275 (0.5525)          & $^{e}$                    \\
e: \revision{HypVINN} (Ours)                         & & 0.7224 (0.1120)          & $^{}$                 & & 0.8683 (0.1074)          & $^{}$        & & 1.4315 (1.7678)          & $^{}$                    \\ \hline
\multicolumn{10}{l}{\textbf{Multi-modal (MM) input (T1w \& T2w)}} \\
f: \revision{MM-VINN} \citep{henschel2022fastsurfervinn}                         & & 0.7918 (0.0924)          & $^{c,d,e}$            & & 0.9033 (0.0774)          & $^{c,e}$     & & 1.1350 (0.5819)          & $^{c,d,e}$   \\
g: \revision{HypVINN} (Ours)                         & & 0.7936 (0.0956)          & $^{b,c,d,e}$          & & 0.9068 (0.0743)          & $^{c,d,e,f}$ & & 1.1207 (0.5563)          & $^{c,d,e}$                \\ \hline
\end{tabular}}
\end{table*}

 This section is divided into four parts with the aim to thoroughly validate our hetero-modal method for hypothalamic sub-regions and adjacent structures segmentation \revision{(referred to as \textit{HypVINN}). The \textit{HypVINN} model is composed of the \textit{HM-VINN} architecture and learning strategies introduced in Sections~\ref{sec:hetero_learning} and~\ref{sec:model_learning}}. i)~Initially, we evaluate the segmentation \textbf{accuracy} of \revision{\mbox{\textit{HypVINN}'s}} predictions against manual annotations. For this purpose, we benchmark the network based on the performance in the unseen test-set against multi- and uni-modal models -- including the only other contemporary method for hypothalamus parcellation (Section~\ref{sec:uni-multi-hetero}), and manual rater variability (Section~\ref{sec:intra-rater}). ii) We assess the \textbf{generalizability} of our method to a different image resolution -- \SI{1.0}{\milli\meter} isotropic MRI scans (Section~\ref{sec:generalizability}). iii) We
test the \textbf{reliability} of the predicted volumes in a  within-session test-retest scenario (Section~\ref{sec:test-set}). iv) Finally, we measure the \textbf{sensitivity} of the proposed pipeline to replicate known hypothalamic volume effects with respect to age and sex. In order to ensure that all experiments are carried out under the same testing \rev{conditions, all} inference analyses are evaluated in a \rev{Docker} container with a 12 GB NVIDIA Titan V GPU. Model inference can also run on the CPU at reduced speeds.

\subsection{Accuracy} \label{sec:accuracy}
In this section, we benchmark and evaluate the accuracy of \revision{the hetero-modal \textit{HypVINN}}. All implemented networks are trained using the scheme mentioned in Section~\ref{sec:model_learning}. 

To show a proof-of-concept for our proposed \revision{\textit{HypVINN}} in segmenting hypothalamic sub-regions and adjacent structures with missing input modalities, we benchmark our method against segmentation scenarios where all modalities are always available (i.e.\ uni-modal and multi-modal models). For this purpose, we implement the classic \revision{\textit{VINN}} with three different inputs: i) only T1w (\revision{\textit{T1-VINN}}), ii) only T2w (\revision{\textit{T2-VINN}}), and iii) T1w \& T2w (\textit{multi-modal \revision{(MM)-VINN}}). For the multi-modal model, the input passed to the network consists of a multi-channel image created by stacking T1w and T2w image slices on top of each other. Additionally, we compare our \revision{\textit{HypVINN}} against the method proposed by \textit{Billot et al.}~\citep{billot2020automated} -- a \textit{3D-UNet} with extensive data augmentation for hypothalamic sub-segmentation on T1w images. Direct comparison of our predicted outcomes with the results from the already trained \revision{model from \textit{Billot et al.}} is not possible as our annotation protocol segments more structures and uses a different hypothalamic parcellation. Therefore, we utilize the implementation provided by the authors to retrain their T1w model from scratch with our manual annotations. It is important to notice that we don't fine-tune \revision{the implementation from \textit{Billot et al.}}, and any optimization of their tool is outside this paper's scope. Furthermore, all comparative \revision{\textit{VINN}} baselines follow the same 2.5D scheme as mentioned in Section~\ref{sec:network}, and inference in \revision{\textit{HypVINN}} is done per input combination. The difference between results in the following two sections is in the data used for training: For Section \ref{sec:uni-multi-hetero} and Table \ref{tab:accuracy_cv_test_set}, all networks are trained in a 4-fold cross-validation scheme to also generate validation performance on the holdout validation split (see~\ref{sec_appendix:ablation} for ablation results). For all other results we used the full training set (n=44). Finally, performance is assessed on the unseen test-set by the three metrics (Dice, HD95, and VS).  

\subsubsection{Comparison with the state-of-the-art} \label{sec:uni-multi-hetero}
In Table~\ref{tab:accuracy_cv_test_set}, we present the similarity scores for the global segmentation performance of all evaluation metrics as well as significance indicators (corrected $p < 0.05$). Here, we observe that \revision{\textit{HypVINN}} performs as well as the modality-specific models. In the T1w-only inference scenario, the \revision{\textit{T1-VINN}} outperforms \revision{\textit{HypVINN}} in Dice and HD95; however, there is no statistical difference between them. On the other hand, when T1w and T2w are available, \revision{\textit{HypVINN}} outperforms the multi-modal model in all evaluation metrics with statistical significance in VS. Furthermore, inputting only a T2w yields the lowest segmentation results from all benchmark models, and the T2w specialized network outranks the \revision{\textit{HypVINN}} with statistical significance. Additionally, we observe that for \revision{\textit{HypVINN}} the inclusion of both modalities improves segmentation performance compared to its single modality counterparts with statistical significance for all metrics in T2w and for T1w only in Dice. For the modality-specific models, \revision{\textit{MM-VINN} and \textit{T1-VINN}}  perform equally well with no statistical significance between them. Finally, our models (both T1 and multi-modal variants) outperform the T1 \textit{3D-UNet} in our segmentation task with statistical significance.

 \revision{We additionally observe that the global results are not driven by any particular structure, as the per-structure results from \textit{HypVINN} and the comparison models align with their respective global outcomes. Furthermore, using a T2w scan as the only source for inferring information is consistently underperforming, both at the global and per-structure levels. For detailed per-structure performance results, see Appendix Figure~\ref{fig_appendix:per_struc_results}}. 
 
 Moreover, the contribution of T2-derived features can also be visualized in \revision{\mbox{\textit{HypVINN}'s}}  learned global fusion weights where the T2-block weight ($\approx 0.25$) has a much lower value than the T1-block weight ($\approx 0.75$) starting already in early stages of training in all implemented networks as shown in Appendix Figure~\ref{fig_appendix:mod_weight_contribution}. Thus,  performance is mainly driven by the T1-derived information, with T2w being only a support modality. For this reason, in the remaining experiments, we only use a T2w image in combination with a T1w image and not as a standalone modality.

\subsubsection{Intra-rater reproducibility}  \label{sec:intra-rater}

In this section, we compare the performance of the automated methods against our main rater variability (i.e.\ intra-rater variability). The intra-rater variability puts the accuracy results into context, where it can be seen as the ideal automated method performance. We assess this variability by computing the similarity between the two sets of manual segmentations of the main rater in the in-house test-set. Note, in contrast to Section \ref{sec:accuracy}, all models are retrained on the full training dataset. It is important to note, that the testing-set is still unseen for these models and is only used for final performance. These "final" models are additionally used for the generalizability (Section~\ref{sec:generalizability}), reliability (Section~\ref{sec:test-set}), and sensitivity (Section~\ref{sec:age-sex}) analyses.

\begin{figure}[!hbt]
    \centering
    \includegraphics[width=0.98\columnwidth]{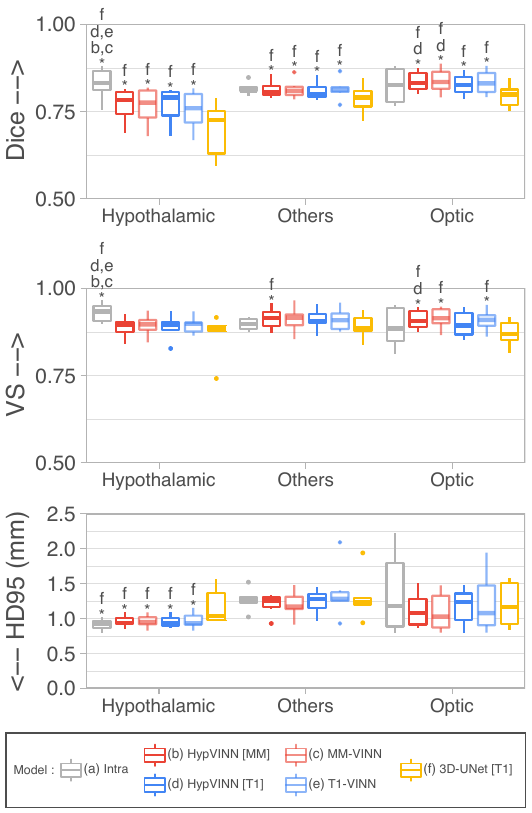}
     \caption{Segmentation performance comparison on the in-house test-set between manual intra-rater scores vs.\ our proposed \revision{\textit{HypVINN}} and benchmark F-CNNs. \revision{\textit{HypVINN}} (dark red and dark blue) produces comparable results to the manual intra-rater agreement (gray). 
     Note: similarity scores are presented for the hypothalamic, others, and optic regions. \revision{Additionally, a letter directly on top of a box plot indicates which other models the model significantly outperforms (paired Wilcoxon signed-rank test, corrected $p < 0.05$}).}
    \label{fig:intra_fsvinn}
\end{figure}

\begin{figure*}[!hbt]
    \centering
    \includegraphics[width=0.95\textwidth]{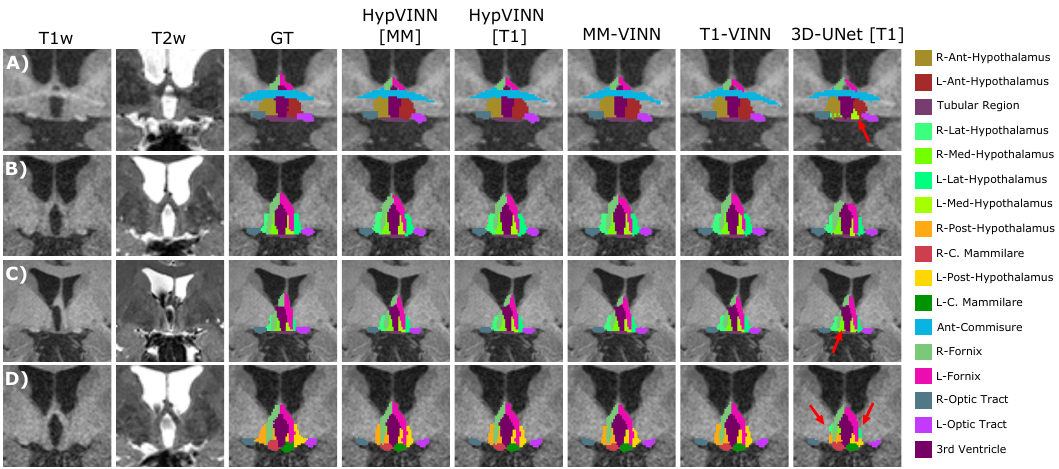}
    \caption{Comparison of the ground truth vs.\ predictions from the proposed \revision{\textit{HypVINN}} and comparison baselines for four participants of the in-house test-set. A-D) All automated methods generate similar segmentation to the manual ones. However, differences are observed in the delineation of the hypothalamic contour. Furthermore, the \textit{3D-UNet} presents the least smooth transitions between hypothalamic structures from all automated methods (red arrows). Note: each row represents a different participant with corresponding MRI modalities (T1-weighted (T1w) and T2w-weighted (T2w)), manual ground truth (GT), and automated generated segmentations on the coronal view. The color scheme for the visible structures is presented on the right.}
    \label{fig:inter_fs_vinn_examples}
\end{figure*}

In Figure~\ref{fig:intra_fsvinn}, we present \revision{box plots} for the three accuracy metrics (Dice, VS, and HD95) in the test-set for the three major regions (hypothalamic, optic, and others, see Section~\ref{sec:labels}). We observe that our main rater has an overall good intra-rater agreement between annotation sessions (Global $\rightarrow$ Dice = 0.8210, VS = 0.9100, HD95 = \SI{1.1277}{\milli\meter}). Furthermore, all automated 2.5D methods perform equally well (Global $\rightarrow$ \revision{\textit{T1-VINN}}: Dice = 0.7869, VS = 0.9017, HD95 = \SI{1.1638}{\milli\meter}; \revision{\textit{MM-VINN}}: Dice = 0.7937, VS = 0.9036, HD95 = \SI{1.0723}{\milli\meter}; \revision{\textit{HypVINN}} with T1 input: Dice = 0.7905, VS = 0.8980, HD95 = \SI{1.1103}{\milli\meter}; \revision{\textit{HypVINN}} with MM input: Dice = 0.7950, VS = 0.9008, HD95 = \SI{1.0857}{\milli\meter}). Additionally, the \textit{3D-UNet} presents the lowest segmentation performance from all final models (Global $\rightarrow$ Dice = 0.7435, VS = 0.8763, HD95 = \SI{1.2347}{\milli\meter}).

The intra-rater scores outperform all the implemented automated methods in Dice and VS, with significant statistical differences present in the hypothalamic region structures (corrected $p < 0.01$). Moreover, the HD95 inter-rater hypothalamic region results are significantly better than the ones of the 3D model. On the other hand, \revision{\textit{MM-VINN} and \textit{HypVINN}} outperforms the intra-rater results in recognizing tissue boundaries (HD95), even if no statistical significance can be inferred from the statistical test. We additionally observe that manually replicating the boundary outline in the structures from the others and optic regions is more challenging. Furthermore, we visually notice that all automated methods generate similar predictions to the manual ones, with the most considerable discrepancies in identifying the hypothalamus contour (outside boundaries), as illustrated in Figure~\ref{fig:inter_fs_vinn_examples}. Moreover, the 3D model generates the noisiest hypothalamic edges from all implemented methods.

Finally, when comparing accuracy results between 2.5D automated methods statistically significant differences are only present in Dice and VS for the optic region between \revision{\textit{HypVINN}} inference setups (corrected $p < 0.05$) with the multi-modal input variation having a better performance (Dice: \textbf{0.8329} vs.\ 0.8238 and VS: \textbf{0.9119} vs.\ 9021). Nonetheless, we also observe improvements without statistical significance in hypothalamic region localization (Dice) and boundary detection (HD95) in structures from the others and optic regions. These results follow the previous section (\ref{sec:uni-multi-hetero}), where \revision{\textit{HypVINN}} shows better segmentation results when all modalities are available. Moreover, the T1 and multi-modal 2.5D counterparts outperform the 3D model, aligning with previous findings. 

\subsection{Generalizability} \label{sec:generalizability}

 In this section, we evaluate the robustness of the proposed hetero-modal model (\revision{\textit{HypVINN}}) to generalize to brain MRI scans with a different image resolution (\SI{1.0}{\milli\meter} isotropic) than the training one (\SI{0.8}{\milli\meter} isotropic). For this purpose, we utilize the MRI scans from the Rhineland Study (RS) in-house test-set (n=6) and a random subset (n=9) of the UK Biobank (UKB) dataset that is manually annotated (see Section~\ref{sec:mri_data}). \revision{For the Rhineland Study, as the MR scans and respective ground truth are at \SI{0.8}{\milli\meter} isotropic resolution, we down-sample the pre-registered T1w and T2w scans from their native resolution to the desired \SI{1.0}{\milli\meter} isotropic resolution. After the 1mm scans are processed by the segmentation model, the resulting probability maps (i.e.\ soft-labels) are up-sampled to the original \SI{0.8}{\milli\meter} resolution. Thereafter, hard labels are generated. This strategy prevents the downsampling of manual labels to \SI{1.0}{\milli\meter}, which introduces interpolation artefacts that could potentially decrease accuracy along boundaries, thereby impacting the analysis.} On the other hand, no resampling is needed for the UK Biobank scans as this dataset is acquired and labeled at \SI{1.0}{\milli\meter} resolution. However, multi-modal evaluation is not done for this dataset as T2w scans are not available. Therefore, we limit the generalizability analysis in the UK Biobank dataset to the performance of the standalone T1w input models. Finally, generalizability performance is assessed by the three similarity metrics (Dice, HD95, and VS) at the native resolution of the corresponding manual reference, except for volume similarity (VS) in the \SI{1.0}{\milli\meter} Rhineland Study predictions. VS does not require spatial overlap between label maps, thus, can be computed without the need for resampling to the same resolution. 

\begin{figure}[!hbt]
    \centering
    \includegraphics[width=0.98\columnwidth]{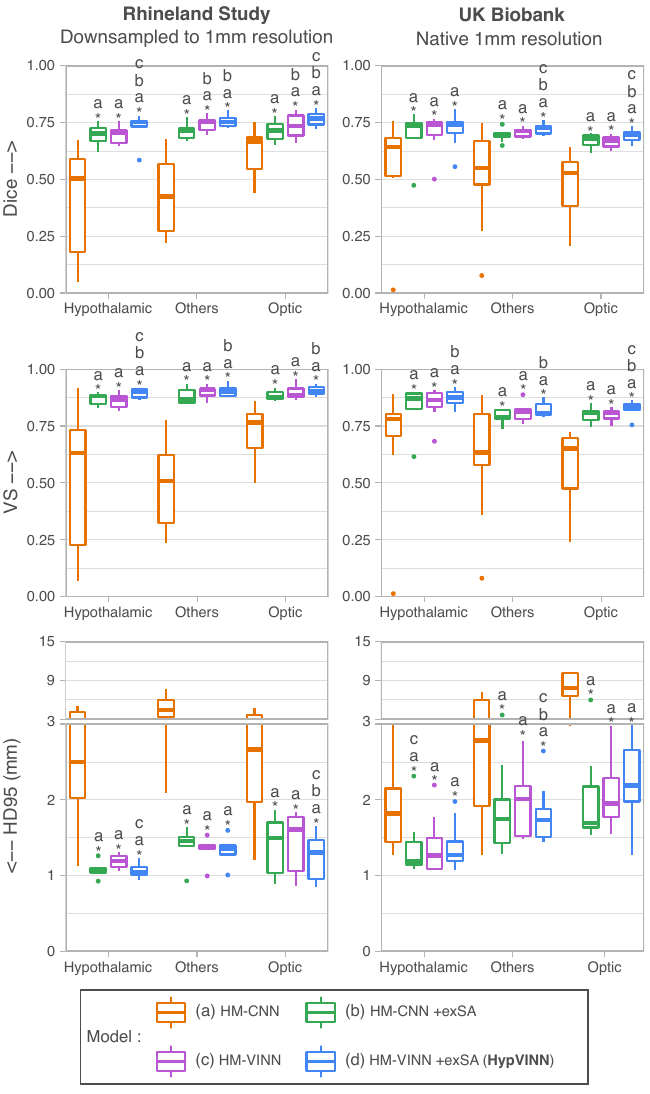}
    \caption{ Retrospectively benchmarking of single resolution (\SI{0.8}{\milli\meter}) trained networks to segment \SI{1.0}{\milli\meter} T1w MR scans from the Rhineland Study and UK Biobank. Our proposed approach \revision{(\textit{HypVINN}) consisting of the \textit{HM-VINN} architecture plus external scale augmentation (+exSA, blue) outperforms other comparison baselines in both manually labeled datasets}. Note: similarity scores are presented for the hypothalamic, others, and optic regions. \revision{Additionally, a letter directly on top of a box plot indicates which other models the model significantly outperforms (paired Wilcoxon signed-rank test, corrected $p < 0.05$).}}
    \label{fig:generilability_res}
\end{figure}

\textit{Henschel et al.}~\citep{henschel2022fastsurfervinn} demonstrated generalizability of \textit{VINN}, \textit{HM-VINN}'s parent architecture, to unseen resolutions. Their results, however, were achieved \rev{training with multi-resolution data}, which is a different scenario to ours, where only \SI{0.8}{\milli\meter} data is available. Therefore, here we further compare generalizability of our \textit{HM-VINN} architecture to segment \SI{1.0}{\milli\meter} MR scans to F-CNNs without resolution-independence mechanisms (\textit{HM-CNN}). \rev{In \textit{HM-CNN}, we replace} the flexible network-integrated resolution-normalization step inside \textit{HM-VINN} with a fixed scale transition. \rev{Furthermore, to isolate the contributions of the proposed resolution-independence scheme, we train both \textit{HM-VINN} and \textit{HM-CNN}  with and without external scale augmentation (exSA). It is important to note that the \textit{HM-VINN} +exSA (proposed \textit{HypVINN}) used in this analysis is the one trained in Section~\ref{sec:intra-rater}. Therefore, to ensure a fair comparison, all benchmarked networks are trained using the same procedure.} We limited this analysis to only T1 input models as T1 is the primary MRI sequence for our segmentation task. Finally, in order to validate the robustness of \revision{\textit{HypVINN}} in both inference scenarios, we compare our method against the modality-specific model from the previous section (i.e.\ \revision{\textit{T1-VINN}, \textit{MM-VINN}} and \textit{3D-UNet}).

\begin{figure}[!t]
    \centering
    \includegraphics[width=0.98\columnwidth]{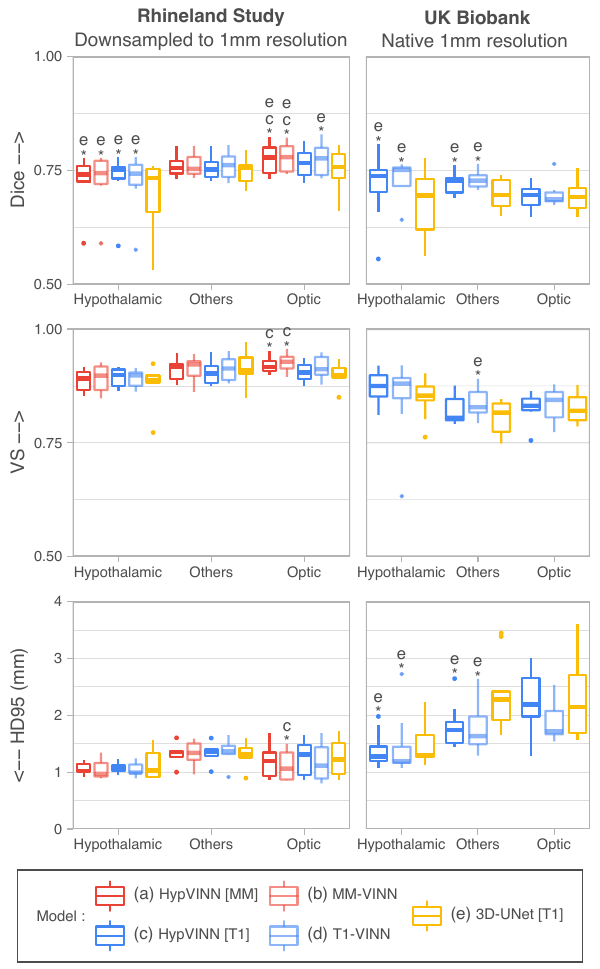}
    \caption{Segmentation performance comparison between our proposed \revision{\textit{HypVINN}}, with multi-modal input (MM) and uni-modal T1 input (T1), vs.\ modality-specific models for segmenting \SI{1.0}{\milli\meter} MR scans from the Rhineland Study and UK Biobank. \revision{\textit{HypVINN}} (dark red and dark blue) can generalize remarkably well to \SI{1.0}{\milli\meter} MR scans independent of the provided MRI input. Note: similarity scores are presented for the hypothalamic, others, and optic regions. \revision{Additionally, a letter directly on top of a box plot indicates which other models the model significantly outperforms (paired Wilcoxon signed-rank test, corrected $p < 0.05$).}}
    \label{fig:generilability_mod}
\end{figure}

\begin{figure*}[!t]
    \centering
    \includegraphics[width=0.95\textwidth]{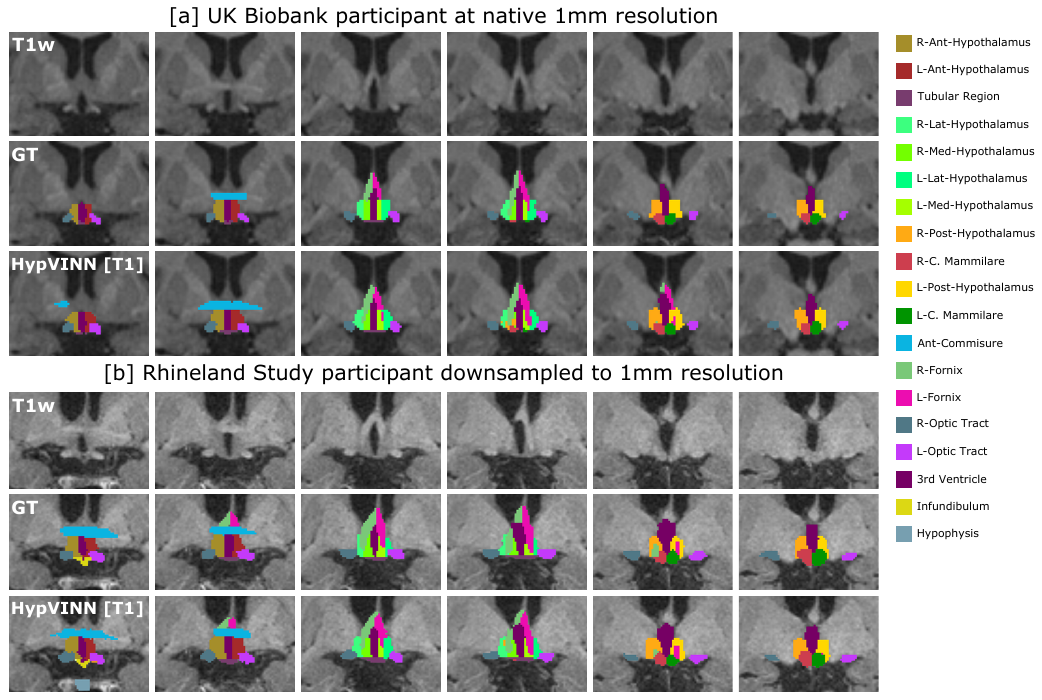}
    \caption{Segmentation examples on the coronal view from our proposed \revision{\textit{HypVINN}} with T1 input and manual ground truth (GT) for one labeled \SI{1.0}{\milli\meter} scan from the UKBiobank and one \SI{1.0}{\milli\meter} scan from the Rhineland Study unseen test-set. Even though our proposed method is not trained with \SI{1.0}{\milli\meter} scans, it can generate accurate predictions at this resolution. Note: the color scheme for the visible structures is presented on the right.}
    \label{fig:generilability_images}
\end{figure*}

In Figures~\ref{fig:generilability_res} and ~\ref{fig:generilability_mod}, we present the generalizability results for the segmentation evaluation metrics in the hypothalamic, optic, and others regions for both datasets. For the first comparison analysis (Figure~\ref{fig:generilability_res}), \rev{the inclusion during training of exSA in both \textit{HM-VINN}  (proposed \textit{HypVINN}, Figure~\ref{fig:generilability_res} blue) and \textit{HM-CNN} (Figure~\ref{fig:generilability_res} green) architectures shows better segmentation performance compared to their respective comparative baseline without exSA (Figure~\ref{fig:generilability_res} orange and purple). Furthermore, we observe that the proposed \textit{HypVINN} (\textit{HM-VINN} +exSA) yields the best segmentation scores among all benchmark networks across different regions and metrics for both datasets, except for HD95 in the optic and hypothalamic structures for UKB. However, the differences in HD95 performance between our \textit{HypVINN} and the \textit{HM-VINN} (optic region) and \textit{HM-CNN} +exSA (hypothalamic region) baselines in the UKB dataset are not statistically significant (corrected $p > 0.1$).} Lastly, as expected, \revision{the vanilla \textit{HM-CNN} (no exSA or resolution-independence)} fails in both datasets for all regions, showcasing the expected generalizability issues of a standalone F-CNN to out-of-distribution resolutions.

Analyzing the generalizability results between input modalities, we observed that even though models have not been trained at \SI{1.0}{\milli\meter} resolution, they can generalize remarkably well, as illustrated in Figure~\ref{fig:generilability_mod} and ~\ref{fig:generilability_images}. For RS, no significant differences are found between 2.5D models except for the optic area where both multi-modal models outperform the T1-input \revision{\textit{HypVINN}} with statistical significance (corrected $p < 0.02$;  metric significance: Dice and VS both methods, and HD95 only \revision{\textit{MM-VINN}}). \rev{In UKB scans, the T1-input \textit{HypVINN} outperforms the T1-specialized model in all metrics for the hypothalamic region. On the other hand, \textit{T1-VINN} outranks our hetero-modal model in the others and optic regions. However, none of the above differences are statistically significant (corrected $p > 0.1$)}. Finally, when comparing against the \textit{3D-UNet} (which has been trained with external scale augmentation), the 2.5D models show in RS significantly better Dice scores for the hypothalamic and optic regions (corrected $p < 0.02$). For UKB, the 2.5D models significantly outperform the \textit{3D-UNet} in Dice and HD95 for the hypothalamic and others regions (corrected $p < 0.01$).

\subsection{Test-retest reliability}  \label{sec:test-set}
Assuming that brain anatomy does not change within the same MR session, a reliable method should generate the same (or very similar) volume estimates from repeated in-session scans acquired under the same conditions (e.g.\ machine, acquisition protocol, region of interest). Here, we benchmark and evaluate the reliability of our proposed hetero-modal F-CNN to predict hypothalamic sub-regions and adjacent structure volumes in a test-rest scenario. For this purpose, we process the T1w and T2w scans from the test-retest dataset (n=21) not only with \revision{\textit{HypVINN}} but also with the benchmark models used in the previous sections (see Sections~\ref{sec:intra-rater} and~\ref{sec:generalizability}) except for the \textit{3D-UNet} as it is the model with the lowest segmentation accuracy results. Since the test-retest dataset includes two T1w scans per participant and only a single T2w scan, the T2w is independently registered two times, each time using a different T1w as reference. Afterwards, we assess the reliability of the methods by computing volume similarity (VS) and intra-class correlation (ICC) between volume predictions across sequences. Finally, we compare the methods' volume similarity performance with a paired two-sided Wilcoxon signed-rank test. 

All methods have an excellent agreement (ICC(A,1) $> 0.95$) between volume predictions across sequences for all regions, as can be seen in  Appendix \rev{Table}~\ref{tab_appendix:icc_vs_fsvinn}. Furthermore, all implemented methods perform equally well for VS in all regions (VS $> 0.98$). Finally, we observe a statistically significant difference in the structures from the others region between \revision{\textit{HypVINN}} with multi-modal input  and \revision{\textit{T1-VINN}} (VS: \textbf{0.9960} vs.\ 9927, corrected $p < 0.05$).  

\subsection{Sensitivity to age and sex effects} \label{sec:age-sex}

\begin{figure*}[!hbt]
    \centering
    \includegraphics[width=0.95\textwidth]{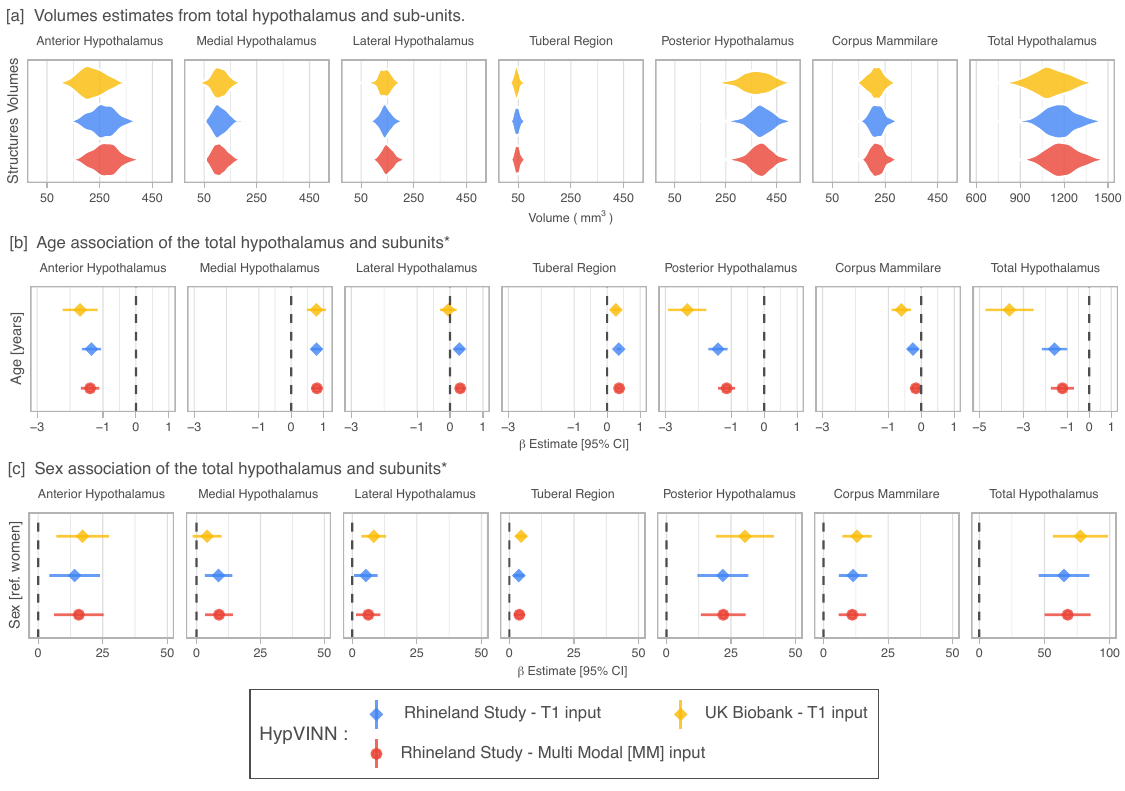}
    \caption{Hypothalamic volumes estimates (a) and volume associations with age (b) and sex (c) in participants from \revision{the Rhineland Study (n=457) and UK Biobank (N=520) for \textit{HypVINN}}. Age and sex effects on hypothalamic volume estimates in the Rhineland Study from \revision{\textit{HypVINN}}, independent of the provided MRI input, follow the same direction trend. Furthermore, our model replicates previous sex findings \revision{in both datasets} corroborating the stability and sensitivity of our method. Note: \textsuperscript{*}Effects are obtained after accounting for head-size (eTIV) and modality sequence \revision{(only Rhineland Study)}.}
    \label{fig:rs_effects}
\end{figure*}

Previous studies have shown that men have a larger hypothalamus volume than women at a global level ~\citep{isiklar2022hypothalamic} but also at a sub-unit level~\citep{makris2013volumetric,thomas2019higher}. Therefore, in this section, we aim to use the automated hypothalamic volume estimates to replicate these findings and explore volume-age correlations in a general population, representing a feasible scenario in which our method will be used as the post-processing analysis pipeline. \revision{To this end, we process the T1w scans from the Rhineland Study (n=463) and UK Biobank (n=535) case-study datasets (see Section~\ref{sec:mri_data}) with our proposed \textit{HypVINN}. To further evaluate the robustness of our hetero-modal model to handle different modalities, we also assess the effects in the Rhineland cases when both pre-registered T1w and T2w scans are available at inference. Ideally, the direction of the effects should not be modified by the input scenarios (only T1w or T1w \& T2w). We note that joint T1w \& T2w analysis in the UK Biobank is not possible due to the absence of T2w scans.}

All generated predictions are visually inspected by an experienced rater. \revision{A total of six participants \rev{(1.29\%)} from the Rhineland Study (RS) and fifteen participants \rev{(2.80\%)} from the UK Biobank (UKB) are excluded from the analysis sample due to segmentation errors (see Appendix Figure~\ref{fig_appendix:rs_failed_examples} for examples). For the remaining participants (RS: n=457, UKB: n=520)}, bias field correction is performed for all T1w and T2w scans as a pre-processing step, and structure volume estimates are compensated for partial volume effects using  \revision{\textit{FastSurfer}'s optimized Python re-implementation of \textit{FreeSurfer}'s mri\_segstats command (\href{https://github.com/Deep-MI/FastSurfer/blob/8d6e7ee4ee24dc24754ece97a10b76745315eddb/FastSurferCNN/segstats.py}{segstats.py})}. Finally, for the total hypothalamus as well as for each of the hypothalamic sub-regions, we calculate the association \revision{per dataset} of age and sex with the respective volumes using independent multivariable linear regression models. \revision{All models are further adjusted for head-size (estimated total intracranial volume, eTIV), and RS models are also corrected for the T1w sequence version ($T1_{seq}$), and T2w sequence version ($T2_{seq}$). Furthermore, de-meaned versions of age ($\hat{age}$) and eTIV ($\hat{eTIV}$) are used in the association analysis (UKB-Model: $Volume \sim \hat{age} + sex + \hat{eTIV}$, RS-Model: $Volume \sim \hat{age} + sex + \hat{eTIV} + T1_{seq} + T2_{seq}$)}.  All statistical analyses are performed in R~\citep{R}, and eTIV estimations are computed using \textit{FreeSurfer}~\citep{buckner2004unified}. It is important to note, that automated segmentations can be carried out without needing bias field corrected scans. Here, we correct the bias field in a pre-processing step primarily for the partial volume estimation, \revision{which is a post-processing step to the segmentation}.

The predicted volumes for the total hypothalamus follow the results from smaller studies~\citep{bocchetta2015detailed,makris2013volumetric,chen2019volume,rodrigues2022benchmark,schindler2013development} with a similar global anatomical definition (from $910~mm^{3}$ to $1580~mm^{3}$) as can been seen in Figure~\ref{fig:rs_effects}~a). \revision{ For the sub-regions, we observe that the tubular region is the smallest segmented hypothalamic structure ($\pm 45.9~mm^{3}$) and the posterior hypothalamus the biggest one ($\pm 379.3~mm^{3}$)}. However, a direct comparison in size of our hypothalamic sub-regions with other studies is not possible due to the different segmentation protocols. 

\begin{figure*}[!hbt]
    \centering
    \includegraphics[width=0.95\textwidth]{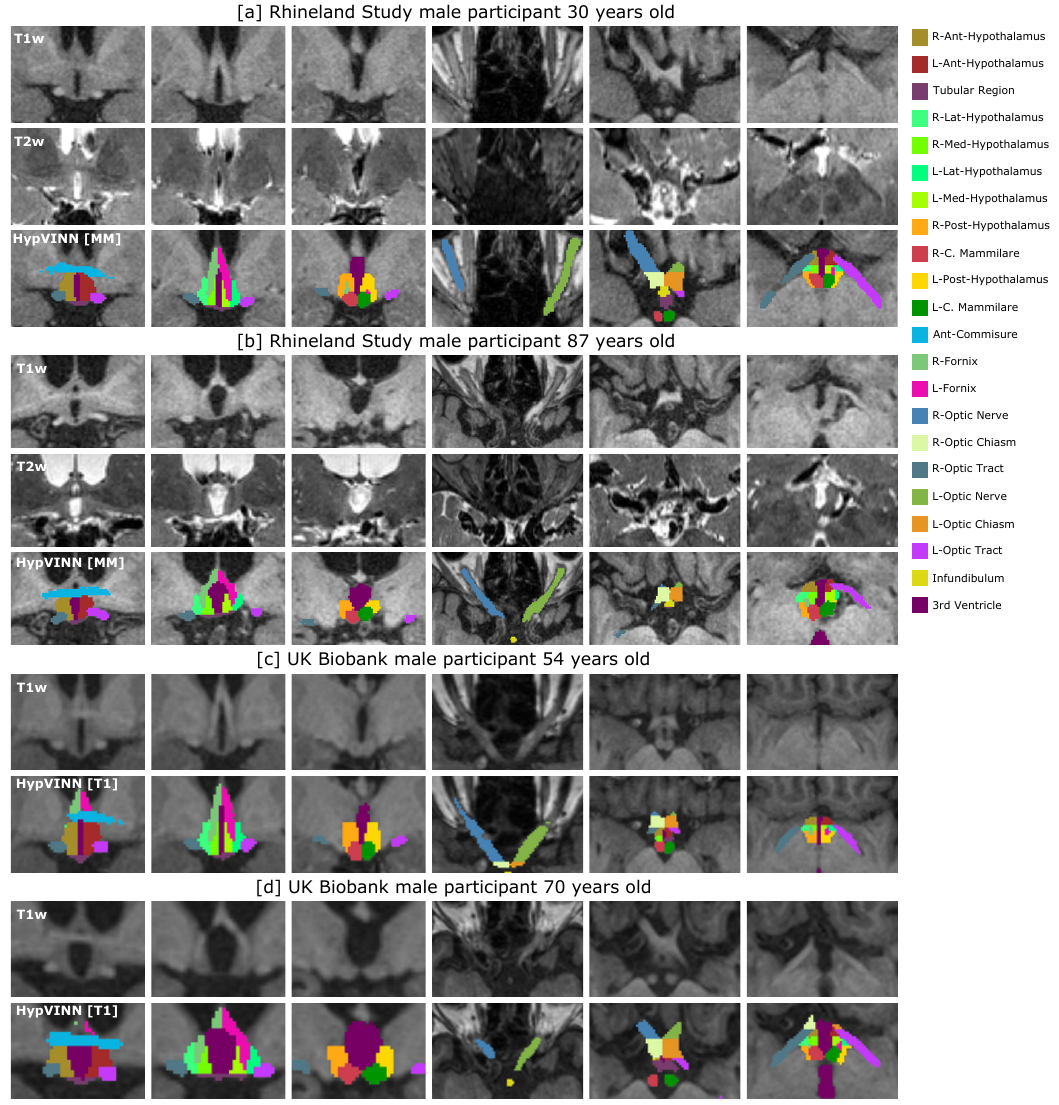}
    \caption{Examples of correct predictions in the Rhineland Study \revision{(a-b) and Uk Biobank (c-d) from our proposed \textit{HypVINN} with multi-modal [MM] or T1w only [T1] input for four unseen random male participants with different ages}. Note: for each participant, T1w, T2w (only Rhineland Study participants), and \revision{\textit{HypVINN}} outcomes are presented. Furthermore, in each participant's row, the first three images display the different hypothalamic structures on the coronal view, and the remaining three images show all remaining structures on the axial view. The color lookup table for all visible structures is presented on the right.}
    \label{fig:rs_examples}
\end{figure*}

\revision{For both RS and UKB subsets}, the total hypothalamus volumes significantly decreased ($p < 0.001$) with age (see  Figure~\ref{fig:rs_effects}~b). This negative association is also observed in the sub-regions except for the middle structures (e.g.\ tuberal-region, medial and lateral hypothalamus), where the volumes are positively correlated with age. \revision{ However, this positive correlation in all middle structures is not observed in the UKB, where a significant increase is not found for the lateral hypothalamus. Furthermore, all structures independent of the dataset, except for the medial hypothalamus in UKB},  show statistically significant sex differences ($p < 0.05$) even after correcting for head-size, with men having larger hypothalamic volumes than women (see Figure~\ref{fig:rs_effects}~c). These results are in line with previous findings~\citep{makris2013volumetric,thomas2019higher,isiklar2022hypothalamic}. Moreover, as expected, all inferred volumes are positively associated with eTIV ($p < 0.01$). 

 \revision{Independent of the provided MRI input, age and sex effects on hypothalamic volume estimates in the Rhineland Study using our \textit{HypVINN} exhibit the same directional trends. Moreover, even though \textit{HypVINN}  is trained with all RS sequence versions}, we observe differences between sequences; however, none of them are significant ($p > 0.05$). Nevertheless, controlling for MRI sequences in any downstream statistical analysis is recommended when including image biomarkers obtained from multiple MRI sequences.

\revision{From the visual quality assessment, we observe that our tool performed very well in two different datasets; examples of correct segmentations for four random male participants with different ages can be observed in Figure~\ref{fig:rs_examples}}. For the failing cases, we note that segmentation errors are mainly present when there is an unclear boundary of the hypothalamus due to severe enlargements of the third ventricle as illustrated in Appendix Figure~\ref{fig_appendix:rs_failed_examples}.

\section{Discussion}  

In this paper, we present the first hetero-modal model for automated sub-segmentation of the hypothalamus and adjacent structures on T1w and T2w brain MRI at isotropic \SI{0.8}{\milli\meter} or \SI{1}{\milli\meter} resolutions. The proposed model can generate accurate segmentations of the twenty-four different structures in less than a minute from a standalone T1w image or by including an additional co-registered T2w image, without requiring multiple input-specific models, thus providing a robust, quick, and reliable solution for assessing hypothalamic volumes in small and large cohorts. 

 Firstly, we introduce a different segmentation protocol of the hypothalamus compared to the one proposed by \textit{Makris et al.}~\citep{makris2013volumetric}. Therefore, we re-train the only other contemporary method for hypothalamus sub-segmentation of \SI{1}{\milli\meter} T1w images~\citep{billot2020automated}. The parcellation method of \textit{Makris et al.} was developed for in-vivo semi-automatic hypothalamic segmentation using 1.5T isotropic \SI{1}{\milli\meter} MR images and was therefore necessarily less detailed than the one presented in this work. In general, we define the boundaries of the hypothalamus as a whole according to the same anatomical definitions and landmarks used by them. Yet, for sub-segmentation of the different hypothalamic subregions, we use a more fine-grained approach to take optimal advantage of the higher spatial resolution offered by the available 3T \SI{0.8}{\milli\meter} isotropic MR images. Consequently, our approach results in the sub-segmentation of more hypothalamic structures as detailed in Table~\ref{tab:gt_labels}. For example, whereas both the posterior hypothalamus and mammillary bodies were included under the label "posterior hypothalamus" in the parcellation scheme of \textit{Makris et al.}, our method provides separate volumetric estimates for each of these structures, which is of clinical relevance given that these structures operate in a functionally independent manner. Another noteworthy difference between the two parcellation schemes concerns the subdivision of the medial part of the hypothalamus: in contrast to \textit{Makris et al.} who subdivided this region into a superior and an inferior tuberal region, we follow the more conventional neuroanatomical subdivision of this region into the medial and the lateral hypothalamus - using the fornix as the boundary between these two structures -, and tubular region. For the tubular region, we group the tuberomammillary region, the median eminence, and the arcuate nucleus. Again, we opt for this approach to gain more detailed anatomical information about the various substructures of the hypothalamus. In addition, our method also provides automatic segmentation of several other important structures in the vicinity of the hypothalamus, for which, until now, no automated segmentation procedure has been available. Notably, these adjacent hypothalamic structures include the hypophysis (i.e.\ the pituitary gland), which is the body's principal and most versatile endocrine gland responsible for the central regulation of most other endocrine tissues throughout the body; the epiphysis, the site where the "sleep hormone" melatonin is synthesized; as well as all major structures of the central optic system including the optic nerves, the optic chiasm, and the optic tracts.
 
 Despite the small size of different sub-structures and low contrast on MR images, our novel deep-learning technique \revision{(\textit{HypVINN})} can accurately segment all twenty-four structures even when input modalities are missing at inference time. \revision{\textit{HypVINN}} performs as well as state-of-the-art modality-specific F-CNNs. Passing a T2w scan as standalone input to \revision{\textit{HypVINN}} or to a specialized T2w model generates the lowest performance from all input variations (see section~\ref{sec:uni-multi-hetero}). For our hetero-modal model, the difference in contribution between T1- and T2-derived information is quantifiable in the modality weights from the fusion module, with the weight of the T1-block ($W_{T1}$) tripling the T2 one. Thus, an available T1w scan is more important for the current segmentation task than a T2w scan. Nonetheless, we demonstrate that including a T2 can still be beneficial for some structures as models with multi-modal information yield generally better segmentation performance.

Unequal performance between inference setups (i.e.\ available input modalities) was also reported in other hetero-modal deep-learning segmentation tasks, with higher results achieved when the primary modality was available  ~\citep{hemis_havaei2016,hetero_variational_dorent2019,pimms_varsavsky2018}. In our case, preference for the T1 modality could be explained by the inherent modality bias from the manual annotation process. Our labeling protocol is mainly performed on the T1w scans, and the T2w scans are only used as a support modality as most anatomical boundaries are visible in T1. Hence, evaluating segmentation performance with the current manual labels is not entirely neutral across the various inference configurations. A more fair evaluation will require training and validation using manual annotations explicitly tailored to a structure's visible anatomical characteristics in each input combination. However, generating $2^{m}-1$ manual labels per participant, where $m$ represents the number of modalities, is not feasible as creating manual annotations for a single configuration is already expensive and time-consuming. Therefore, based on our findings, we recommend utilizing a T2w scan accompanied by a T1w scan (i.e.\ multi-modal input) and not as a standalone input for the current segmentation task.

Our hetero-modal model, when including a T1w image, exhibits segmentation performance in the range of the main rater variability (see Section~\ref{sec:intra-rater}). The intra-rater variability can be seen as the ideal performance of the automated method as we use manually annotated labels from the main rater to train our F-CNNs. Therefore, it is challenging for an automated approach to outperform the intra-rater scores. Considering this, the accuracy in the hypothalamic region of our hetero-modal model and all benchmark methods is lower than the intra-rater agreement on all evaluation metrics. Yet, the underperformance in this region can also be attributed to the low MR contrast between neighboring structures, especially for the medial and lateral hypothalamus. Nonetheless, the segmentation results are en-par with other deep-learning techniques on similar brain segmentation tasks (i.e.\ small size and low contrast across anatomical boundaries)~\citep{billot2020automated,estrada2021automated}.

\revision{\textit{HypVINN}} not only performs well on segmenting isotropic \SI{0.8}{\milli\meter} T1w and T2w MR scans, but it also exhibits generalizability to isotropic \SI{1}{\milli\meter} MR scans from the Rhineland Study and UK Biobank dataset (see Section~\ref{sec:generalizability}). We demonstrate that utilizing the resolution-independence mechanism performs as well as external scale augmentations to handle unseen resolution when training with a single (\SI{0.8}{\milli\meter}) resolution. Furthermore, we show that resolution-independence  combined with external scale augmentations \revision{(proposed)} outperforms all other comparative baselines. 

Furthermore, \revision{\textit{HypVINN}} performs equally well as modality specific-models in both \SI{1}{\milli\meter} datasets. 
As expected, performance on the Rhineland Study data is higher than on the UK Biobank. The UK Biobank dataset, consists of scans from a different cohort and is acquired with a different MRI acquisition protocol. Due to these dissimilarities, segmentation performance is not directly comparable. Nevertheless, the proposed \revision{\textit{HypVINN}} generalizes quite well to this external dataset. Finally, even though our model supports both \SI{0.8}{\milli\meter} and \SI{1}{\milli\meter} resolutions, we recommend to process \SI{0.8}{\milli\meter} MR scans at their native resolution to obtain more detailed and precise predictions by leveraging the additional information present in the higher resolution.  \revision{Note, our proposed model also shows promising results in the high-resolutional MRI scans from the Human Connectome Project (HCP) young adult and lifespan pilot project datasets~\citep{hcp-van2012human,bookheimer2019lifespan,harms2018extending}; see Appendix Figure~\ref{fig_appendix:hcp_examples} for prediction examples of our tool in HCP scans.}

Throughout this work, we compare our \revision{\textit{HypVINN}} against the re-trained version of the \textit{3D-UNet} with extensive data augmentations proposed by \textit{Billot et al.}~\citep{billot2020automated} for hypothalamus sub-segmentation. Our results demonstrate that our method not only outperforms the \textit{3D-UNet} in terms of segmentation accuracy (see Sections~\ref{sec:uni-multi-hetero} and \ref{sec:intra-rater}) but also exhibits better generalizability across both comparative datasets (see Section~\ref{sec:generalizability}). Additionally, the training process for the \textit{3D-UNet} using the authors' released implementation and recommended training parameters takes approximately 100 hours per model using the GPU setup described in Section~\ref{sec:model_learning}. In contrast, back-to-back training of the three F-CNNs that compose our \revision{\textit{HypVINN}} takes around 19 hours (roughly 6 hours per F-CNN). Therefore, besides outperforming the contemporary method, our approach can be (re)trained more efficiently with a lower carbon footprint.

 As demonstrated in the Rhineland Study data, all automated methods exhibit excellent test-retest agreement between in-session volume estimates (see Section~\ref{sec:test-set}). \revision{Additionally, our \textit{HypVINN} shows high robustness and generalizability across the general population of the Rhineland Study and UK Biobank case-study datasets, with only twenty-one cases (2.10\%) between the two datasets being excluded from the age and sex analysis due to segmentation errors (see Section~\ref{sec:age-sex})}. The most common factor for our pipeline to fail is a severe deformation of the third ventricle (i.e.\ out-of-distribution cases), which generates unclear hypothalamic boundaries, as illustrated in Appendix Figure~\ref{fig_appendix:rs_failed_examples}. \revision{ Therefore, careful inspection is recommended when using our tool in aging populations and clinical cohorts, as the prevalence of large ventricles increases with age and certain diseases (e.g.\ Alzheimer's disease, Parkinson's disease, etc.)}. We recommend visually inspecting the predictions \revision{from scans with pathological changes and from} volumetric outliers within the cohort before including them in any downstream analysis, particularly outliers from the third ventricle and medial/lateral hypothalamus. Although volumetric outlier detection can help identify predictions with significant failures, more robust quality control tools are desirable. However, developing these tools is outside this paper's scope and will be future work.

In line with previous studies on smaller datasets~\citep{makris2013volumetric,thomas2019higher,isiklar2022hypothalamic}, we also find that the volume of the total hypothalamus is larger in men compared to women. \revision{However, our analyses in two substantially larger population-based cohorts revealed that the volumes of virtually all hypothalamic substructures are significantly larger in men independent of head size}. Our findings thus warrant further detailed association studies to investigate the clinical relevance of these pronounced sex differences in the human hypothalamus. On the other hand, the derived age effects from small-scale studies present inconsistent results for the different hypothalamic substructures, except for the total hypothalamus whose total volume decreases with age~\citep{makris2013volumetric,bocchetta2015detailed,billot2020automated,isiklar2022hypothalamic}. Our method's total hypothalamic volume estimates also replicate this negative correlation with age. Furthermore, although most hypothalamic regions atrophy with increasing age, the volume of the middle/tuberal region of the hypothalamus significantly increases with age. This finding is novel and could imply that specific hypothalamus regions could be resistant to age-associated atrophy. Indeed, the paraventricular nucleus contained within the medial hypothalamic region exhibits a striking stability in terms of neuronal numbers, both with age and in the context of common neurodegenerative diseases such as Alzheimer's disease~\citep{lucassen1994activation}. These findings thus underscore the need for further large-scale studies into the differential effects of age on different hypothalamic substructures. 

In conclusion, we demonstrate that \revision{\textit{HypVINN}} can successfully identify the desired structures with similar or better performance than state-of-the-art modality-specific models regarding segmentation accuracy, generalizability, and test-retest reliability. \revision{ Furthermore, the fact that \textit{HypVINN} replicates previous age and sex findings on large unseen subsets of the Rhineland Study and the UK Biobank corroborates the stability and sensitivity of our method. Moreover, our hypothalamic sub-segmentation tool} generates accurate segmentations regardless of whether both T1w and T2w images are available or just a single T1w image. However, utilizing both modalities results in slightly improved segmentation outcomes.

\rev{Future work will focus on supporting a wider range of resolution by training our \textit{HypVINN} with multi-resolution, thus fully exploiting the advantages of using a voxel-size independent F-CNN (VINN)~\citep{henschel2022fastsurfervinn}. Moreover, we will also focus on improving the robustness of our tool to out-of-distribution cases (e.g.\ severe deformation of the third ventricle). Since \textit{HypVINN} is based on deep learning, boosting the robustness to these cases can potentially be achieved by retraining with manual annotations created on participants with low segmentation quality or by applying realistic non-linear deformations as an additional data augmentation during the training process~\citep{faber2022cerebnet}. Finally, extending the input flexibility of our tool to scenarios where input scans are at different resolutions (mixed resolutions) is also of interest, as it could allow the deployment of our tool in more scenarios where HighRes data is unavailable in all modalities.}  

Overall, we introduce \revision{\textit{HypVINN}} -- the first hetero-modal deep learning method for hypothalamic sub-segmentation and segmentation of other adjacent structures, such as the hypophysis, epiphysis, and major structures of the central optic system. The proposed method offers a more detailed parcellation of the hypothalamus compared to the only other contemporary automated method~\citep{billot2020automated}. Additionally, it can generate accurate segmentations from T1w and T2w MR images at isotropic \SI{0.8}{\milli\meter} or \SI{1}{\milli\meter} resolutions. \revision{Finally, \textit{HypVINN} will be incorporated into the \textit{FastSurfer} neuroimaging software suite. Thus, providing an easy to use alternative for more reliable assessment of hypothalamic imaging-derived phenotypes.} 

\section*{Data and code availability statement}

This work uses MRI data from the Rhineland Study and UK Biobank. The Rhineland Study data is not publicly available because of data protection regulations. However, access can be provided to scientists in accordance with the Rhineland Study’s Data Use and Access Policy. Requests to access the data should be directed to Dr. Monique Breteler at RS-DUAC@dzne.de.  UK Biobank data are available through a procedure described at http://www.ukbiobank.ac.uk/using-the-resource/.

The method presented in this article will be made publicly available on Github (https://github.com/Deep-MI/FastSurfer) upon acceptance.

\section*{Credit authorship contribution statement}
\textbf{Santiago Estrada}: Methodology, Software, Validation, Formal analysis, Investigation, Conceptualization, Writing - original draft, Writing - review \& editing, Visualization.
\textbf{David K\"ugler}: Conceptualization, Methodology, Validation, Data Curation, Writing - original draft, Writing - review \& editing.
\textbf{Emad Bahrami}: Investigation, Software, Validation, Data Curation, Writing - original draft.
\textbf{Peng Xu} : Data Curation, Investigation, Validation, Writing - original draft.
\textbf{Dilshad Mousa}: Data Curation, Investigation.
\textbf{Monique M.B.\ Breteler}: Supervision, Funding acquisition, Resources, Writing - review \& editing.
\textbf{N.\ Ahmad Aziz}: Conceptualization, Validation, Resources, Writing - original draft, Writing - review \& editing, Supervision, Funding acquisition.
\textbf{Martin Reuter}: Conceptualization, Validation, Resources, Writing - original draft, Writing - review \& editing, Supervision, Project administration, Funding acquisition.

\section*{Declaration of interests}

The authors declare that they have no conflict of interest.

\section*{Ethics statement}
This work uses MRI data from participants of the Rhineland Study and UK Biobank. Participants in both studies gave written informed consent in accordance with the ethical guidelines of the individual studies. The  Rhineland study is carried out in accordance with the recommendations of the International Council for Harmonisation (ICH) Good Clinical Practice (GCP) standards (ICH-GCP). UK Biobank had obtained ethics approval from the North West Multicentre Research Ethics Committee.

\section*{Acknowledgment}
We would like to thank the Rhineland Study group for supporting the data acquisition and management. This work was supported by DZNE institutional funds, by the Federal Ministry of Education and Research of Germany (031L0206, 01GQ1801), the Chan Zuckerberg Initiative (Project FastSurfer, \rev{Grant Number: EOSS5 2022-252594}), the Helmholtz-AI project DeGen (ZT-I-PF-5-078), by an Alzheimer's Association Research Grant (Award Number: AARG-19-616534), and by NIH (R01 LM012719, R01 AG064027, R56 MH121426, and P41 EB030006). Peng Xu was supported by a scholarship from China Scholarship Council and N.Ahmad Aziz was supported by a European Research Council Starting Grant (Number: 101041677). 

\revision{This research has been conducted using the UK Biobank Resource under Application Number 82056. Data in the appendix were also provided in part by  the publicly available Human Connectome Project (HPC), WU-Minn Consortium (Principal Investigators: David Van Essen and Kamil Ugurbil; 1U54MH091657) funded by the 16 NIH Institutes and Centers that support the NIH Blueprint for Neuroscience Research; and by the McDonnell Center for Systems Neuroscience at Washington University.}

\newpage
\onecolumn
\appendix
\section{}
\setcounter{figure}{0}
\renewcommand{\thefigure}{A\arabic{figure}}
\setcounter{table}{0}
\renewcommand{\thetable}{A\arabic{table}}

\begin{table*}[!hbt]
\centering
\caption{Sequence parameters for the T1-weighted and T2-weighted versions in the Rhineland Study. To date, there have been two versions of the T1w sequence ($T1w^{a-b}$) and four versions of the T2w sequence ($T2w^{a-d}$) - care was taken to preserve the image contrast between versions for both sequences.}
\label{tab:t1_t2_parameters}
\resizebox{0.95\textwidth}{!}{%
\renewcommand{\arraystretch}{1.5} % Default value: 1
\begin{threeparttable}
\begin{tabular}{llccclllccccclc}
\hline
\multicolumn{5}{c}{T1w sequence}                                      &  & \multicolumn{9}{c}{T2w sequence}                                                                                                                                                        \\ \cline{1-5} \cline{7-15} 
\multirow{2}{*}{Parameters} &  & \multicolumn{3}{c}{Version}          &  & \multirow{2}{*}{Parameters} &  & \multicolumn{7}{c}{Version}                                                                                                                            \\ \cline{3-5} \cline{9-15} 
                            &  & $T1w^{a}$         &  & $T1w^{b}$               &  &                             &  & $T2w^{a}$             &                      & $T2w^{b}$                 &                      & $T2w^{c}$             &  & $T2w^{d}$                 \\ \cline{1-5} \cline{7-15} 
Repetion  time (TR)         &  & \multicolumn{3}{c}{\SI{2560}{\milli\second}}          &  & Repetion  time (TR)         &  & \multicolumn{7}{c}{\SI{2800}{\milli\second}}                                                                                                                            \\
Inversion time (TI)         &  & \multicolumn{3}{c}{\SI{1100}{\milli\second}}          &  & Echo time (TE)              &  & \multicolumn{7}{c}{\SI{405}{\milli\second}}                                                                                                                             \\
Flip angle                  &  & \multicolumn{3}{c}{7$^{\circ}$}                &  & Matrix size                 &  & \multicolumn{7}{c}{$320 \times 320 \times 224$}                                                                                                                    \\
Matrix size                 &  & \multicolumn{3}{c}{$320 \times 320 \times 224$}  &  & Phase-encoding direc.\textsuperscript{++}       &  & A\textgreater{}P &                      & R\textgreater{}L     &                      & A\textgreater{}P &                      & A\textgreater{}P     \\
PI acc. factor              &  & 1x3 &  & 1x2                 &  & PI acc. factor              &  & \multicolumn{3}{c}{3x1}                                        &                      & 2x1              &                      & 1x2\textsuperscript{+}                 \\
Readout bandwith            &  & 240 Hz/pixel &  & 740 Hz pixel       &  & PI ref. scan                &  & \multicolumn{3}{c}{Integrated}                                 &                      & \multicolumn{3}{c}{External}                                   \\
Echo time (TE)              &  & \SI{2.94}{\milli\second}\textsuperscript{*}      &  & \SIrange{1.68}{6.51}{\milli\second}\textsuperscript{**} &  & Acquisition  time (TA)      &  & 3:57~\SI{}{\minute}         &                      & 4:30~\SI{}{\minute}             &                      & \multicolumn{3}{c}{4:47 min}                                   \\
Acquisition  time (TA)      &  & 3:43~\SI{}{\minute}    &  & 6:35~\SI{}{\minute}           &  &                             &  &                  & \multicolumn{1}{l}{} & \multicolumn{1}{l}{} & \multicolumn{1}{l}{} &                  &                      & \multicolumn{1}{l}{} \\ \hline
\end{tabular}
\begin{tablenotes}  
\item \textsuperscript{*} 1 echo, \textsuperscript{**} 4 echoes combined to 1.
\item \textsuperscript{+} with one CAIPIRINHA shift~\citep{breuer2006controlled}, \textsuperscript{++} A: anterior, P: posterior, R: right, and L: Left.
\end{tablenotes}
\end{threeparttable}
}
\end{table*}

\begin{table*}[!hbt]
\centering
\caption{Demographics of the Rhineland Study participants for all different datasets. Descriptive data were expressed as mean (SD) or count (percentage) for continuous or categorical variables, respectively. Inter group differences were compared with the Student’s t-test for continuous variables and with the Pearson’s chi-square test for categorical variables.}
\label{tab_appendix:rs_demo_table}
\resizebox{0.8\textwidth}{!}{%
\renewcommand{\arraystretch}{1.2} % Default value: 1
\begin{tabular}{lclclclclcr}
\hline
           & & Case Study   & & Test-Retest  & & In-house    & & Total         & & \multirow{2}{*}{p-value}\\ 
           & & (N=463)      & & (N=21)       & & (N=50)      & & (N=534)       & &    \\ \hline
Sex        & &              & &              & &             & &               & & 0.801\\ 
Women      & & 276 (59.6\%) & & 11 (52.4\%)  & & 30 (60.0\%) & & 317 (59.4\%)  & & \\
Men       & & 187 (40.4\%) & & 6 (47.6\%)   & & 20 (40.0\%) & & 217 (40.6\%)  & & \\ \hline
Age        & &              & &              & &             & &               & & 0.805\\
Mean (SD)  & & 54.9 (14.2)  & & 56.4 (9.3)   & & 54.0 (15.2) & & 54.9 (14.1)   & & \\
Range      & & 30.0 - 95.0  & & 40.0 - 74.0  & & 31.0 - 79.0 & & 30.0 - 95.0   & & \\ \hline
T1w version & &              & &              & &             & &               & & 0.061\\
a          & & 71 (15.3\%)  & & 0 (0.0\%)    & & 4 (8.0\%)   & & 75 (14.0\%)   & & \\
b          & & 392 (84.7\%) & & 21 (100.0\%) & & 46 (92.0\%) & & 459 (86.0\%)  & & \\ \hline
T2w version & &              & &              & &             & &               & & $< 0.001$\\
a          & & 71 (15.3\%)  & & 0 (0.0\%)    & & 4 (8.0\%)   & & 75 (14.0\%)   & & \\
b          & & 14 (3.0\%)   & & 0 (0.0\%)    & & 2 (4.0\%)   & & 16 (3.0\%)    & & \\
c          & & 269 (58.1\%) & & 0 (0.0\%)    & & 27 (54.0\%) & & 296 (55.4\%)  & & \\
d          & & 109 (23.5\%) & & 21 (100.0\%) & & 17 (34.0\%) & & 147 (27.5\%)  & & \\
\hline
\end{tabular}%
}
\end{table*}

\begin{table*}[!ht]
\centering
\caption{Demographics for the training and testing in-house dataset. Descriptive data were expressed as mean (SD) or count (percentage) for continuous or categorical variables, respectively. Inter group differences were compared with the Student’s t-test for continuous variables and with the Pearson’s chi-square test for categorical variables.}
\label{tab_appendix:rs_train_demo_table}
\resizebox{0.9\textwidth}{!}{%
\renewcommand{\arraystretch}{1.2} % Default value: 1
\begin{tabular}{lcllllclclcr}
\hline
            & & \multicolumn{4}{l}{Trainset}                                           & & Testset     & & Total       & & \multirow{2}{*}{p-value}\\ \cline{3-6}
            & &  Split\_1 (N=11) & Split\_2 (N=11) & Split\_3 (N=11) & Split\_4 (N=11) & & (N=6)       & & (N=50)      & &  \\ \hline
Sex         & &                  &                 &                 &                 & &             & &             & & 0.857\\
Women     & & 6 (54.5\%)       & 7 (63.6\%)      & 8 (72.7\%)      & 6 (54.5\%)      & & 3 (50.0\%)  & & 30 (60.0\%) & & \\
Man        & & 5 (45.5\%)       & 4 (36.4\%)      & 3 (27.3\%)      & 5 (45.5\%)      & & 3 (50.0\%)  & & 20 (40.0\%) & & \\ \hline
Age         & &                  &                 &                 &                 & &             & &             & & 0.439\\
Mean (SD)   & & 46.7 (14.8)      & 53.5 (16.0)     & 56.5 (15.3)     & 58.5 (15.0)     & & 55.2 (14.9) & & 54.0 (15.2) & & \\
Range       & & 31.0 - 69.0      & 31.0 - 77.0     & 32.0 - 79.0     & 35.0 - 76.0     & & 35.0 - 71.0 & & 31.0 - 79.0 & & \\ \hline
\end{tabular}%
}
\end{table*}

\begin{table*}[!ht]
\centering
\caption{\revision{Demographics for the UK Biobank participants for all different datasets. Descriptive data were expressed as mean (SD) or count (percentage) for continuous or categorical variables, respectively. Inter group differences were compared with the Student’s t-test for continuous variables and with the Pearson’s chi-square test for categorical variables.}}
\label{tab_appendix:ukb_demo_table}
\resizebox{0.7\textwidth}{!}{%
\renewcommand{\arraystretch}{1.2} 
\begin{tabular}{lclclclcr}
\hline
          & & Case Study         & & Generalizability       & & Total           & & \multirow{2}{*}{p-value}\\ 
          & & (N=535)            & & (N=9)                  & & (N=544)         & & \\ \hline
Sex       & &                    & &                        & &                 & & 0.857\\
Woman     & & 281 (52.5\%)       & & 5 (55.6\%)             & & 286 (52.6\%)    & &  \\
Men       & & 254 (47.5\%)       & & 4 (44.4\%)             & & 258 (47.4\%)    & &  \\ \hline
Age       & &                    & &                        & &                 & & 0.050\\
Mean (SD) & & 63.9 (7.7)         & & 58.7 (11.3)            & & 63.8 (7.8)      & & \\
Range     & & 46.0 - 82.0        & & 45.0 - 77.0            & & 45.0 - 82.0     & & \\ \hline
\end{tabular}

}
\end{table*}

\begin{figure*}[!hbt]
    \centering
    \includegraphics[width=0.95\textwidth]{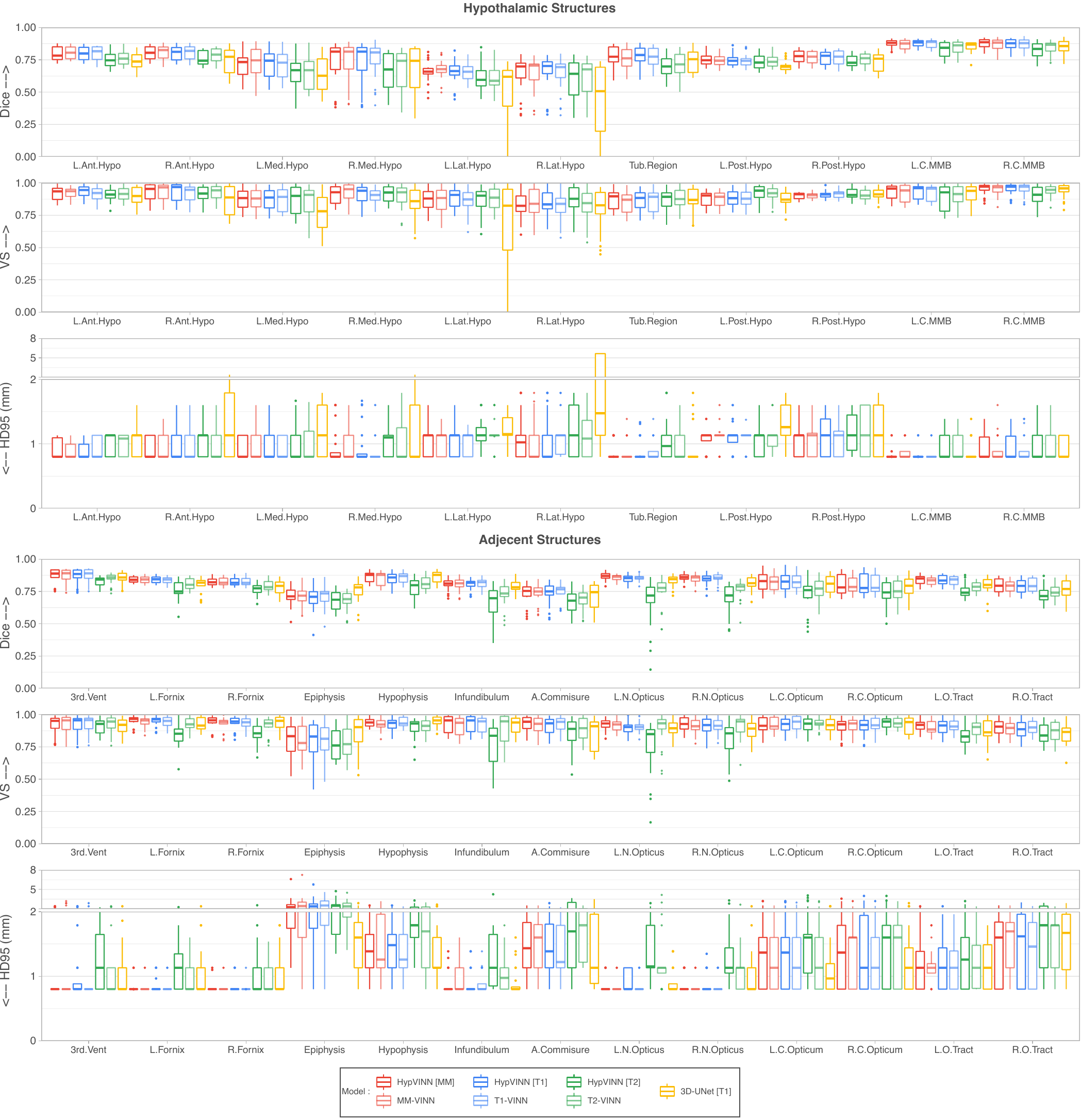}
    \caption{Per structure segmentation performance of the F-CNN models on the unseen in-house test-set. We observe that models with a T1w image as part of its input have comparable results in all structures to the global ones. However, there is a slight decrease in Dice performance in the medial and lateral hypothalamus (Dice $<$ 0.75) compared to the other hypothalamic structures for the 2.5D models. For the 3D model, a similar trend is also observed in the medial hypothalamus; however, in the lateral hypothalamus, performance drastically diminishes in all evaluation metrics (Dice $<$ 0.5, VS $<$ 0.8, and HD95 $>$ \SI{1.2}{\milli\meter}). Furthermore, for the adjacent hypothalamic structures, all 2.5D models present difficulties in localizing the epiphysis and recognizing its boundaries ( Dice $\le$ 0.75, VS $\le$ 0.8, and HD95 $\ge$ \SI{2}{\milli\meter}). Moreover, the epiphysis is the only structure from the twenty-four segmented ones where the 3D model outperforms the T1 and multi-modal comparative baselines (Dice = 0.7558, VS = 0.8571, and HD95 = \SI{1.6386}{\milli\meter}). \revision{Finally, using a T2w scan as the only source for inferring information is consistently underperforming in all structures, especially in the optic region (e.g.\ optic nerve) and middle hypothalamic region (e.g.\ medial and lateral hypothalamus and tubular region). Nonetheless, the inclusion of T2w into the current segmentation task appears to be beneficial as \textit{HypVINN} with multi-modal input outperforms its T1w-only counterpart in most structures (Dice: 16/24, VS: 14/24, and HD95: 18/24).}}
    \label{fig_appendix:per_struc_results}
\end{figure*}

\begin{figure}[!hbt]
    \centering
    \includegraphics[width=0.85\textwidth]{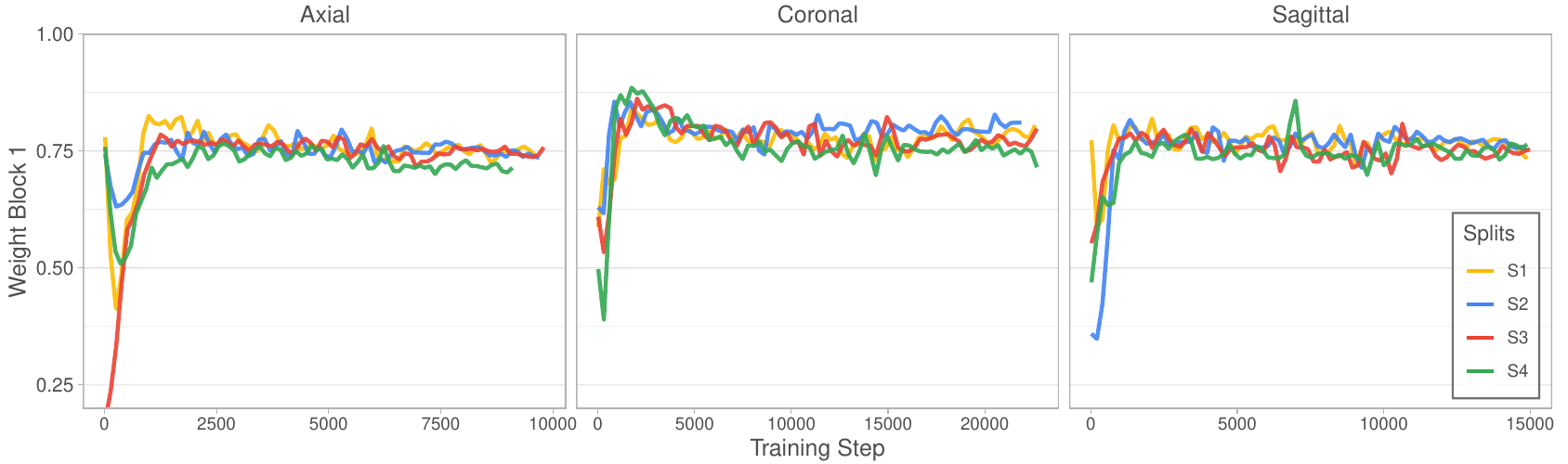}
    \caption{T1-Block learnable modality weight during training. The T1-block has a much higher value ($\approx 0.75$) than the T2-block weight ($\approx 0.25$) in \revision{\textit{HypVINN}'s} fusion module, starting in the early training steps in all four cross-validation training splits (i.e. S1, S2, S3, and S4) . Thus,  performance is mainly driven by the T1-derived information, with T2w being only a support modality.}
    \label{fig_appendix:mod_weight_contribution}
\end{figure}

\begin{table*}[!hbt]
\centering
\caption{\rev{Test-retest reliability: Intra-class correlation (ICC) with a 95\% confidence intervals and volume similarity (VS) between volume estimates across sequences in a test-retest scenario for the 21 cases of the test-retest dataset. All automated methods exhibit excellent test-retest agreement between in-session volume estimates. Note: the statistical significance column (Signif.) indicates, which other models the model outperforms (Wilcoxon signed rank test, corrected $p < 0.05$).}}
\label{tab_appendix:icc_vs_fsvinn}
\resizebox{0.98\textwidth}{!}{%
\renewcommand{\arraystretch}{1.2} % Default value: 1
\begin{tabular}{llllllllllllllll}
\hline
\multirow{3}{*}{Model}  & \multicolumn{5}{c}{Hypothalamic}                          &  & \multicolumn{4}{c}{Others}                             &  & \multicolumn{4}{c}{Optic}                              \\ \cline{3-6} \cline{8-11} \cline{13-16} 
                        &  & ICC (A,1)                 &  & \multicolumn{2}{l}{VS}  &  & ICC (A,1)                 &  & \multicolumn{2}{l}{VS}  &  & ICC (A,1)                 &  & \multicolumn{2}{l}{VS}  \\ \cline{3-3} \cline{5-6} \cline{8-8} \cline{10-11} \cline{13-13} \cline{15-16} 
                        &  & ICC {[}95\% CI{]}         &  & Mean (SD)     & Signif. &  & ICC {[}95\% CI{]}         &  & Mean (SD)     & Signif. &  & ICC {[}95\% CI{]}         &  & Mean (SD)     & Signif. \\ \hline
\textbf{Only T1w input} &  &                           &  &               &         &  &                           &  &               &         &  &                           &  &               &         \\
a: T1-VINN              &  & 0.984 {[}0.959 - 0.994{]} &  & 0.990 (0.011) &         &  & 0.997 {[}0.993 - 0.999{]} &  & 0.993 (0.006) &         &  & 0.982 {[}0.953 - 0.993{]} &  & 0.994 (0.005) &         \\
b: HypVINN (Ours)       &  & 0.982 {[}0.953 - 0.993{]} &  & 0.987 (0.025) &         &  & 0.999 {[}0.997 - 1.000{]} &  & 0.996 (0.003) &         &  & 0.985 {[}0.955 - 0.994{]} &  & 0.994 (0.005) &         \\ \hline
\multicolumn{16}{l}{\textbf{Multi-modal (MM) input (T1w \& T2w)}}                                                                                                                                           \\
c: MM-VINN              &  & 0.990 {[}0.975 - 0.996{]} &  & 0.990 (0.010) &         &  & 0.998 {[}0.995 - 0.999{]} &  & 0.994 (0.006) &         &  & 0.972 {[}0.879 - 0.990{]} &  & 0.992 (0.006) &         \\
d: HypVINN (Ours)       &  & 0.984 {[}0.957 - 0.994{]} &  & 0.989 (0.015) &         &  & 0.999 {[}0.998 - 1.000{]} &  & 0.996 (0.003) &  $^{a}$       &  & 0.986 {[}0.955 - 0.995{]} &  & 0.994 (0.004) &         \\ \hline
\end{tabular}}
\end{table*}

\begin{figure}[!ht]
    \centering
    \includegraphics[width=0.9\textwidth]{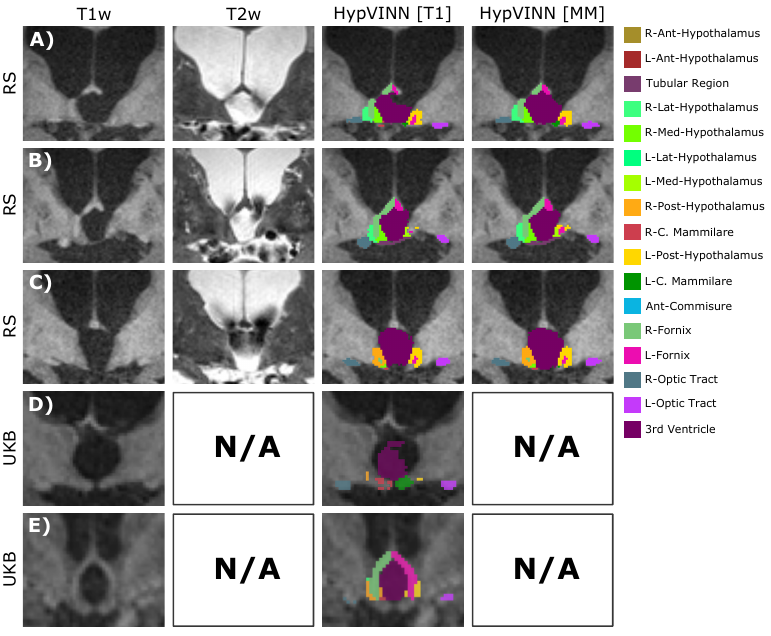}
    \caption{\revision{Examples of excluded cases  from the Rhineland Study (RS) and UK Biobank (UKB) after visual quality assessment. A) to E)} Unclear boundary of the hypothalamus due to severe enlargements of the third ventricle (i.e.\ out-of-distribution cases) producing segmentation errors. Note: each row represents a different participant with corresponding MRI modalities (T1-weighted (T1w) and \revision{T2w-weighted (T2w) -- if available)}, and automated generated segmentations on the coronal view. The color scheme for the visible structures is presented on the right.}
    \label{fig_appendix:rs_failed_examples}
\end{figure}

\begin{figure}[!ht]
    \centering
    \includegraphics[width=0.98\textwidth]{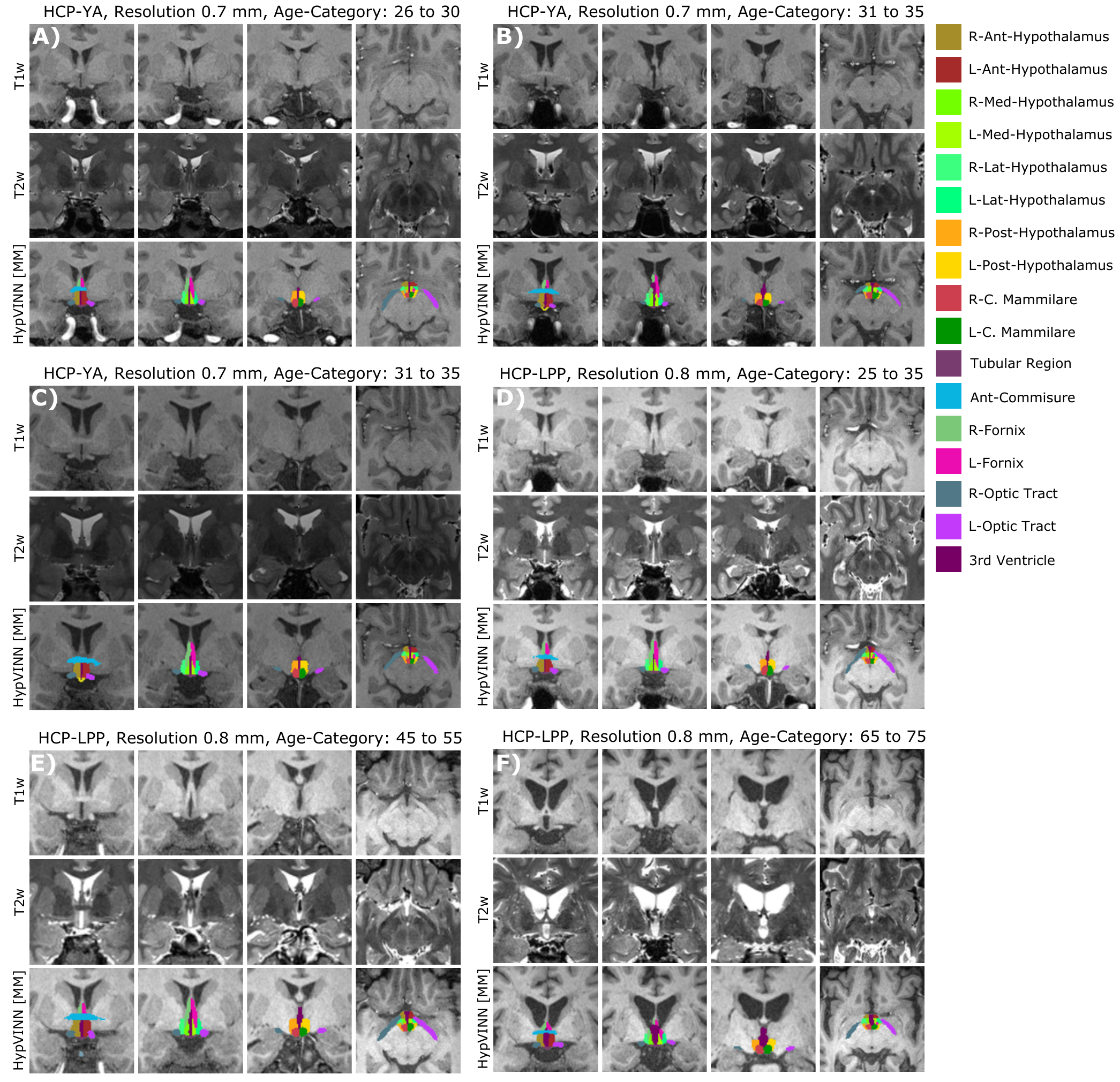}
    \caption{\revision{Examples of correct predictions in the Human Connectome Project (HCP) young adults (HCP-YA, A-C) and HCP lifespan pilot project (HCP-LPP, D-E) datasets~\citep{hcp-van2012human,bookheimer2019lifespan,harms2018extending} from our proposed \textit{HypVINN} with multi-modal input (MM) for six random participants. We observe that our tool shows promising results in both available HCP resolutions (\SI{0.7}{\milli\meter} and \SI{0.8}{\milli\meter}). Furthermore, our tool seems to generalize well across age categories inside the training age range (training data started at age 30). However, all the above observations are only qualitative, and no accuracy segmentation metrics can be computed as manual annotations are unavailable for this dataset. Note: T1w, T2w, and \textit{HypVINN} outcomes are presented for each participant. Furthermore, in each participant's row, the first three images display the different hypothalamic structures on the coronal view, and the remaining image shows the structures on the axial view. The color lookup table for all visible structures is presented on the right.}}
    \label{fig_appendix:hcp_examples}
\end{figure}

\clearpage
\section{Ablation analysis} \label{sec_appendix:ablation}

\setcounter{figure}{0}
\renewcommand{\thefigure}{B\arabic{figure}}
\setcounter{table}{0}
\renewcommand{\thetable}{B\arabic{table}}

We execute ablation analysis to optimize the fusion module weighting scheme inside the \revision{\textit{HM-VINN}} architecture by training the model with global and per-channel modality weights. First, all networks are trained from scratch using the four data-splits from the in-house training-set in a leave-one-out cross-validation approach. Afterwards, the best model is chosen based on the cross-validation performance in the hold-out validation-sets. The three evaluation metrics (Dice, VS, HD95) are computed per input modality combination (i.e.\ only T1w or only T2w, or both)  between the predicted maps after view aggregation and manuals labels. Finally, improvements in segmentation performance are confirmed by statistical testing (corrected $p < 0.05$).

We observe that utilizing global weights outperforms per-channel weights in all comparative metrics and all inference scenarios with statistical significance for the standalone T2w input in all three metrics and for the T1w \& T2w input only in Dice, as presented in Table~\ref{tab_appendix:fusion_fsvinn_ablative}. Therefore, we utilize the global weighting scheme as the fusion module configuration for this work.

\begin{table*}[!ht]
\centering
\caption{\textbf{Fusion module weighting scheme optimization}: Mean (and standard deviation) of segmentation performance metrics per input modality of the ablative \revision{\textit{hetero-modal VINN (HM-VINN)}} architectures on the validation set. Global weights outperforms per channel weights in all comparative metrics and all inference scenarios. Note: the statistical significance column (Signif.) indicates, which other models the model outperforms (Wilcoxon signed rank test, corrected $p < 0.05$).}
\label{tab_appendix:fusion_fsvinn_ablative}
\resizebox{\textwidth}{!}{%
\renewcommand{\arraystretch}{1.2} % Default value: 1
\begin{tabular}{lllllllllll}
\hline
\multicolumn{11}{l}{\textbf{Only T1w input }} \\
\multicolumn{2}{l}{\textbf{Experimental Setup}}      & & \multicolumn{2}{l}{\textbf{Dice} \(\uparrow\)}          & & \multicolumn{2}{l}{\textbf{VS} \(\uparrow\)}            & & \multicolumn{2}{l}{\textbf{HD95 (mm)} \(\downarrow\)} \\ \cline{1-2} \cline{4-5} \cline{7-8} \cline{10-11}
Model       & Weighting Scheme & & Mean(SD)                 & Signif.           & & Mean(SD)                 & Signif.           & & Mean(SD)                 & Signif.      \\ \hline
a: \revision{HM-VINN}  & Global        & & \textbf{0.8068 (0.0841)} &                   & & \textbf{0.9164 (0.0748)} &                   & & \textbf{1.0916 (0.8579)} &              \\
b: \revision{HM-VINN}  & Per Channel   & & 0.8042 (0.0864)          &                   & & 0.9160 (0.0753)          &                   & & 1.0953 (0.7277)          &          \\\hline
\multicolumn{11}{l}{\textbf{Only T2w input}} \\
\multicolumn{2}{l}{\textbf{Experimental Setup}}      & & \multicolumn{2}{l}{\textbf{Dice} \(\uparrow\)}         & & \multicolumn{2}{l}{\textbf{VS} \(\uparrow\)}           & & \multicolumn{2}{l}{\textbf{HD95 (mm)} \(\downarrow\)}     \\ \cline{1-2} \cline{4-5} \cline{7-8} \cline{10-11}
Model       & Weighting Scheme & & Mean(SD)                 & Signif.           & & Mean(SD)                 & Signif.           & & Mean(SD)                & Signif.       \\ \hline 
a: \revision{HM-VINN}  & Global        & & \textbf{0.7354 (0.1115)} & $^{b}$            & & \textbf{0.8753 (0.1166)} & $^{b}$            & & \textbf{1.4154 (1.3291)}& $^{b}$         \\
b: \revision{HM-VINN}  & Per Channel   & & 0.7119 (0.1236)          &                   & & 0.8424 (0.1424)          &                   & & 1.700 (2.3105)          &                 \\ \hline
\multicolumn{11}{l}{\textbf{T1w \& T2w input}} \\
\multicolumn{2}{l}{\textbf{Experimental Setup}}               & & \multicolumn{2}{l}{\textbf{Dice} \(\uparrow\)}           & & \multicolumn{2}{l}{\textbf{VS} \(\uparrow\)}            & & \multicolumn{2}{l}{\textbf{HD95 (mm)} \(\downarrow\)}    \\ \cline{1-2} \cline{4-5} \cline{7-8} \cline{10-11}
Model  & Weighting Scheme      & & Mean(SD)                  & Signif.           & & Mean(SD)                 & Signif.           & & Mean(SD)                 & Signif.        \\ \hline
a: \revision{HM-VINN}  & Global        & & \textbf{0.8128 (0.0814)}  & $^{b}$            & & \textbf{0.9202 (0.0706)} &                   & & \textbf{1.0508 (0.6965)} &                   \\
b: \revision{HM-VINN}  & Per Channel   & & 0.8079 (0.0869)           &                   & & 0.9187 (0.0754)          &                   & & 1.0678 (0.7445)          &                  \\ \hline

\end{tabular}%
}
\end{table*}
\newpage

\clearpage
\section{Criteria for manual annotation of hypothalamic adjacent structures}\label{sec_appendix:manual}
\setcounter{figure}{0}
\renewcommand{\thefigure}{C\arabic{figure}}
\setcounter{table}{0}
\renewcommand{\thetable}{C\arabic{table}}

In Tables \ref{tab_appendix:other_annotations} and \ref{tab_appendix:hypo_annotations}, we present the criteria for manual annotation of hypothalamic adjacent structures and sub-regions in T1w and T2w images. The support of a T2w image was omitted for segmenting UK Biobank data as this data was unavailable. Furthermore, no protocol modification was carried out due to the differences in data resolution -- Rhineland Study \SI{0.8}{\milli\meter} isotropic resolution and UK Biobank \SI{1}{\milli\meter} isotropic resolution.

\begin{table*}[!hbt]
\caption{Criteria for manual annotation of hypothalamic adjacent structures}
\centering
\label{tab_appendix:other_annotations}
\resizebox{\textwidth}{!}{%
\renewcommand{\arraystretch}{1.2}
\begin{threeparttable}
\begin{tabular}{p{0.09\textwidth}p{0.01\textwidth}p{0.05\textwidth}p{0.01\textwidth}p{0.60\textwidth}p{0.01\textwidth}p{0.22\textwidth}}
\hline
\textbf{Structure} & & \textbf{Bilateral\textsuperscript{*}} & & \textbf{Labeling\textsuperscript{**}}     & & \textbf{Note}                        \\ \hline
Optic system  & & Yes & & The optic systems is composed of the optic nerves,tracts and chiasms. The optic chiasm was separated from the optic nerves and tracts at an angle orthogonal to the chiasm at the optic nerve–chiasm and optic tract–chiasm junctions, respectively~\citep{avery2016quantitative}. & & Using axial T1 weighted images.  \\ \hline 

Anterior commissure & & No & & A thick fiber bundle above the $3^{rd}$ ventricle and underneath the anterior horns of the lateral ventricles. It can easily be identified using the brain ventricles and optic tracts as landmarks~\citep{gungor2017white}. & & Labelling on coronal sections in the rostro-caudal direction on T1 weighted images.  \\ \hline 

Fornices & & Yes & & Thick white matter fiber bundles that were labelled in the area where they touch the anterior commissure rostrally and merge with the mammillary bodies caudally; this part of the fornix is generally referred to as the “columna fornicis”. & & Using coronal sections of T1 weighted sequences.  \\ \hline 

Hypophysis  & & \multirow{2}{0.05\textwidth}{No} & & \multirow{2}{0.60\textwidth}{A relatively round structure inferior to the $3^{rd}$ ventricle and rostral to the brain stem, occupying the sella turcica.} & & \multirow{2}{0.22\textwidth}{Using sagittal, axial, and coronal sections of T1 and T2 weighted images.}  \\ 
(i.e. the pituitary gland) & &  & &  & &  \\ \hline 

Infundibulum & & \multirow{2}{0.05\textwidth}{No} & & \multirow{2}{0.60\textwidth}{The stalk-like structure that connects the hypophysis to the hypothalamus.} & &  \\ 
(i.e. the pituitary stalk) & &  & &  & &  \\ \hline

Epiphysis  & & \multirow{3}{0.05\textwidth}{No} & & \multirow{3}{0.60\textwidth}{A low-intensity (on T1 weighted images), pine-shaped unpaired midline brain structure that lies between the caudal recess of the third ventricle and the quadrigeminal cistern~\citep{park2020pineal}.} & & \multirow{3}{0.22\textwidth}{ Labelling was done on coronal sections by moving caudally from the posterior commissure, with its contours demarcated by its pine-like shape and the surrounding cerebrospinal fluid.}  \\ 
(i.e. the pineal gland) & &  & &  & &  \\
& &  & &  & &  \\ 
& &  & &  & &  \\ 
& &  & &  & &  \\ 
& &  & &  & &  \\ \hline

\multirow{5}{0.09\textwidth}{$3^{rd}$ ventricle} & & \multirow{5}{0.05\textwidth}{No} & & Anterior border: lamina terminalis. & & \multirow{5}{0.22\textwidth}{Using sagittal, axial, and coronal sections of T1 and T2 weighted images.} \\
& & & & Lateral border: hypothalamus and thalamus. & &  \\
& & & & Superior border: the roof of the third ventricle starts anteriorly at the foramen of Monro and ends posteriorly in the suprapineal recess. & & \\
& & & & Posterior border:  the posterior commissure, the pineal body, the habenular commissure, and the  suprapineal recess above~\citep{patel2012surgical}. & & \\
& & & & Inferior border: formed from anterior to posterior by the optic recess, the infundibular recess, the tuber cinereum, the mamillary bodies, and the posterior perforated substance~\citep{chaichana2019comprehensive}. & &  \\ \hline      

\end{tabular}
\begin{tablenotes}  
\item \textsuperscript{*} Bilateral structures were defined as those regions that could be separated into a (non-contiguous) left and right half with respect to the midsagittal plane. 
\item \textsuperscript{**}  Labelling was mainly done using T1 weighted images, unless specified otherwise.
\end{tablenotes}
\end{threeparttable}}
\end{table*}
\clearpage
\begin{table*}[!hbt]
\caption{Criteria for manual annotation of hypothalamic sub-regions}
\centering
\label{tab_appendix:hypo_annotations}
\resizebox{\textwidth}{!}{%
\renewcommand{\arraystretch}{1.2}
\begin{threeparttable}
\begin{tabular}{p{0.09\textwidth}p{0.01\textwidth}p{0.05\textwidth}p{0.01\textwidth}p{0.60\textwidth}p{0.01\textwidth}p{0.22\textwidth}}
\hline
\textbf{Structure}  & & \textbf{Bilateral\textsuperscript{*}}      & & \textbf{Labeling\textsuperscript{**}}    & & \textbf{Note}                        \\ \hline
\multirow{6}{0.09\textwidth}{Anterior hypothalamus} & & \multirow{6}{0.05\textwidth}{Yes} & & Medial border: $3^{rd}$ ventricle. & & \multirow{6}{0.22\textwidth}{The \textbf{supraoptic nuclei} were included in this region and were not labelled separately as the spatial resolution was too low for accurate segmentation of these small structures.} \\
& & & & Lateral border: lateral border of the optic tract and the other adjacent white matter tracts~\citep{lemaire2011white}. & & \\
& & & & Anterior border: lamina terminalis attached to the optic chiasm. & & \\
& & & & Posterior border: vanishment of the anterior commissure on coronal sections in the rostro-caudal direction (coinciding with the coronal plane through the posterior border of the anterior commissure and the anterior tip of the infundibulum). & &  \\
& & & & Superior border: horizontal plane through the anterior commissure. & & \\                                                                                
& & & & Inferior border: optic chiasm and infundibulum~\citep{DUDAS202145}  & & \\ \hline 

\multirow{6}{0.09\textwidth}{Medial hypothalamus} & & \multirow{6}{0.05\textwidth}{Yes} & & Medial border: $3^{rd}$ ventricle. & & \\
& & & & Lateral border: fornices. & & \\
& & & & Anterior border: vanishment of the anterior commissure on coronal sections in the rostro-caudal direction.& & \\
& & & & Posterior border: appearance of the mammillary bodies on coronal sections in the rostro-caudal direction. & &  \\
& & & & Superior border: the diencephalic fissure. & & \\                                                                                
& & & & Inferior border: the boundaries of the tuberal region underneath~\citep{makris2013volumetric}.  & & \\ \hline                                                                                                    
\multirow{6}{0.09\textwidth}{Lateral hypothalamus} & & \multirow{6}{0.05\textwidth}{Yes} & & Medial border: fornices. & & \\
& & & & Lateral border: optic tract and the other adjacent white matter tracts. & & \\
& & & & Anterior border: vanishment of the anterior commissure on coronal sections in the rostro-caudal direction.& & \\
& & & & Posterior border: appearance of the mammillary bodies on coronal sections in the rostro-caudal direction. & &  \\
& & & & Superior border: the diencephalic fissure. & & \\                                                                                
& & & & Inferior border: the boundaries of the tuberal region and basal cistern underneath.  & & \\ \hline 

\multirow{6}{0.09\textwidth}{Posterior hypothalamus} & & \multirow{6}{0.05\textwidth}{Yes} & & Medial border: $3^{rd}$ ventricle. & & \\
& & & & Lateral border: white matter tracts. & & \\
& & & & Anterior border: appearance of the mammillary bodies on coronal sections in the rostro-caudal direction. & & \\
& & & & Posterior border: vanishment of the mammillary bodies on coronal sections in the rostro-caudal direction. & &  \\
& & & & Superior border: horizontal plane through the diencephalic fissure. & & \\                                                                                
& & & & Inferior border: boundaries with the mammillary bodies below.  & & \\ \hline      

\multirow{4}{0.09\textwidth}{Tubular region} & & \multirow{4}{0.05\textwidth}{No} & &  The area was defined as the region underneath the $3^{rd}$ ventricle and enclosed by the mammillary bodies caudally and the anterior hypothalamus rostrally, with its superior and inferior borders on each side defined by the horizontal planes going through the superior border of the floor of the third ventricle and the interpeduncular cistern, respectively. & & \\
& & & & Median eminence: The protuberant region between the unpaired infundibular nucleus and the mammillary bodies that had a low-intensity on the sagittal view on T2 weighted sequences. & & \\
& & & & Infundibular nucleus: Dorsocaudal to the junction of the infundibulum (i.e. the pituitary stalk) and the hypothalamus, and was labelled on the sagittal view using T1 (high-intensity) and T2 (low-intensity) weigheted images. & & \\
& & & & Tubero-mammillary nucleus: The remaining areas in the tuberal region. & &  \\ \hline   

Mammillary bodies & & Yes & & Two small, rounded structures at the caudal end of the $3^{rd}$ ventricle. These structures were labelled using both coronal sections in the rostro-caudal direction and axial sections in the dorso-medial direction on T1 weighted images. & & \\ \hline
\end{tabular}
\begin{tablenotes}  
\item \textsuperscript{*} Bilateral structures were defined as those regions that could be separated into a (non-contiguous) left and right half with respect to the midsagittal plane. 
\item \textsuperscript{**}  Labelling was mainly done using T1 weighted images, unless specified otherwise.
\end{tablenotes}
\end{threeparttable}}
\end{table*}

\twocolumn
\bibliographystyle{model5-names}
\biboptions{authoryear,round}

\bibliography{mybibfile.bib}

\end{document}